\documentclass[10pt,twocolumn,letterpaper]{article}

\usepackage[pagenumbers]{cvpr} %

\usepackage{amsmath}
\usepackage[dvipsnames]{xcolor}

\usepackage[inline]{enumitem}
\setlist[itemize]{noitemsep,leftmargin=*,topsep=0em}
\setlist[enumerate]{noitemsep,leftmargin=*,topsep=0em}
\newcommand{\enumlabel}{\textbf{(\roman*)}}
\usepackage[colorinlistoftodos,prependcaption,textsize=footnotesize]{todonotes}

\usepackage{colortbl}
\definecolor{lightgray}{gray}{0.9}

\usepackage[pagebackref,breaklinks,colorlinks]{hyperref}
\usepackage[capitalize]{cleveref}
\crefname{section}{Sec.}{Secs.}
\Crefname{section}{Section}{Sections}
\Crefname{table}{Table}{Tables}
\crefname{table}{Tab.}{Tabs.}
\def\paperID{206} %
\def\confName{3DV\xspace}
\def\confYear{2024\xspace}

\newcommand{\ours}{Lang3DSG\xspace}
\title{Lang3DSG: Language-based contrastive pre-training\\ for 3D Scene Graph prediction}

\author{
Sebastian Koch$^{1,2,3}$\qquad Pedro Hermosilla$^4$\qquad Narunas Vaskevicius$^{1,2}$\qquad \\Mirco Colosi$^2$\qquad Timo Ropinski$^3$ \vspace{0.05cm}\\ 
{\small $^1$\text{Bosch Center for Artificial Intelligence}\quad $^2$Robert Bosch Corporate Research}\\ {\small $^3$\text{University of Ulm} \quad $^4$ TU Vienna}
\\{\small\href{https://kochsebastian.com/sgrec3d}{kochsebastian.com/lang3dsg}}
}

\newcommand{\mytextcolor}[2]{{\scriptsize \textcolor{#1}{#2}}}
\newcommand{\tablepilotstudy}{\begin{table}[t]
\tabcolsep=1.0mm
\footnotesize
\centering
\begin{tabular}{@{}lcccc@{}}
\toprule
       & \multicolumn{2}{c}{Object} & \multicolumn{2}{c}{Predicate}\\
       \cmidrule(lr){2-3}\cmidrule(lr){4-5}\cmidrule(lr){4-5}
       & R@5   $\uparrow$              & mR@5   $\uparrow$             & R@3         $\uparrow$                     & mR@3    $\uparrow$                          \\
      \midrule
      Graph-baseline & 0.63  & 0.30   & 0.94  & 0.57 \\
      STRL~\cite{Choy_2019_CVPR} & 0.75 \mytextcolor{green}{(+0.12)} & 0.35 \mytextcolor{green}{(+0.05)} & 0.94 \mytextcolor{red}{(-0.00)} & 0.50 \mytextcolor{red}{(-0.07)}\\
      DepthContrast~\cite{Zhang_2021_ICCV} & 0.77 \mytextcolor{green}{(+0.14)} & 0.36 \mytextcolor{green}{(+0.06)} & 0.94 \mytextcolor{red}{(-0.00)}  & 0.51 \mytextcolor{red}{(-0.06)}\\
\bottomrule
\end{tabular}
\caption{\textbf{Comparison among point cloud-based 3D scene graphs pre-training.} We conduct a pilot study comparing existing point cloud-based pre-training studies with a graph baseline without pre-training.}
\label{tab:pretrain_study}
\vspace{-0.5cm}
\end{table}
}

\newcommand{\tablemainevaluation}{
\begin{table}[t]
\tabcolsep=2.2mm
\centering
\footnotesize
\begin{tabular}{@{}lcccccc@{}}
\toprule
       & \multicolumn{2}{c}{Object} & \multicolumn{2}{c}{Predicate} & \multicolumn{2}{c}{Relationship} \\
       \cmidrule(lr){2-3}\cmidrule(lr){4-5}\cmidrule(lr){6-7}
      Method & R@5                  & R@10                 & R@3                 & R@5                & R@50                 & R@100                \\
      \midrule
SGGPoint \cite{Zhang_2021_CVPR} & 0.28 & 0.36 & 0.68 & 0.87 & 0.08 & 0.10\\
MSDN \cite{Li_2017_ICCV} & 0.61 & 0.72 & 0.86 & 0.94 & 0.47 & 0.53\\
KERN \cite{Chen_2019_CVPR} & 0.67 & 0.77 & 0.83 & 0.96 & 0.51 & 0.58\\ 
BGNN \cite{Li_2021_CVPR} & 0.71 & 0.82 & 0.87 & 0.94 & 0.55 & 0.60\\
3DSSG \cite{Wald_2020_CVPR} & 0.68                 & 0.78                 & 0.89                & 0.93               & 0.40                 & 0.66                 \\ 
Liu \etal \cite{Liu_2022_TVCG} & 0.74 & 0.83 & 0.90 & 0.96 & 0.62 & 0.68\\
SGFN \cite{Wu_2021_CVPR} & 0.70 & 0.80 & \textbf{0.97} & \textbf{0.99} & 0.85 & 0.87 \\
Ours & \textbf{0.77} & \textbf{0.84} & 0.96 & \textbf{0.99} & \textbf{0.87} & \textbf{0.89} \\
\bottomrule
\end{tabular}
\caption{\textbf{3D scene graph prediction on 3DSSG.} Experimental results for 3D scene graph prediction on 3DSSG. We report the top-k recall values for object classification, predicate prediction as well as relationship prediction. For a fair comparison, all works use ground-truth class-agnostic instance segmentation.}
\label{tab:main_evaluation}
\vspace{-0.4cm}
\end{table}
}

\newcommand{\tablepretrainevaluation}{
\begin{table}[t]
\tabcolsep=2.1mm
\centering
\footnotesize
\begin{tabular}{@{}lcccc@{}}
\toprule
       & \multicolumn{2}{c}{Object} & \multicolumn{2}{c}{Predicate}\\
       \cmidrule(lr){2-3}\cmidrule(lr){4-5}
       & R@5            & mR@5        & R@3                      & mR@3                    \\
      \midrule
    \rowcolor{lightgray}
      Graph-baseline (w/o pre-train) & 0.63  & 0.30   & 0.94  & 0.57 \\
      \rowcolor{lightgray}
      STRL \cite{Choy_2019_CVPR} & 0.75   & 0.35  & 0.94  & 0.50 \\
      \rowcolor{lightgray}
      DepthContrast \cite{Zhang_2021_ICCV} & \textbf{0.77}  & 0.36  & 0.94   & 0.51 \\
      \midrule
      \midrule
      Ours (no-graph) & 0.74 & 0.37 & 0.94 & 0.60\\
      Ours & \textbf{0.77} & \textbf{0.43} & \textbf{0.96} & \textbf{0.67}\\
\bottomrule
\end{tabular}
\caption{\textbf{Pre-training comparison.} A simple graph baseline outperforms existing point cloud-based pre-training methods on predicate prediction. Our novel pre-training shows high effectiveness outperforming existing pre-training approaches and graph baseline.\looseness=-1}
\label{tab:pretrain_evaluation}
\vspace{-0.45cm}
\end{table}
}

\newcommand{\tablepretrainsupervision}{
\begin{table}[t]
\tabcolsep=4.0mm
\centering
\footnotesize
\begin{tabular}{@{}lcccc@{}}
\toprule
       & \multicolumn{2}{c}{Object} & \multicolumn{2}{c}{Predicate}\\
       \cmidrule(lr){2-3}\cmidrule(lr){4-5}
       & R@5 & mR@5  & R@3  & mR@3           \\
      \midrule
      Relationship  & 0.75 & 0.38 & 0.94 & 0.63 \\
      Object + Predicate  & 0.77 & 0.41 & \textbf{0.96} & 0.66 \\
      Obj + Pred + Rel  & \textbf{0.77} & \textbf{0.43} & \textbf{0.96} & \textbf{0.67} \\
\bottomrule
\end{tabular}
\caption{\textbf{Pre-training language supervision.} We ablate what language supervision is required for 3D scene graph pre-training. Combining relationship supervision with separate object and predicate supervision yields the best results.}
\label{tab:pretrain_supervision}
\vspace{-0.4cm}
\end{table}
}

\newcommand{\tableablationcbertclip}{
\begin{table}[t]
\tabcolsep=3.4mm
\centering
\footnotesize
\begin{tabular}{@{}lcccc@{}}
\toprule
       & \multicolumn{2}{c}{Object} & \multicolumn{2}{c}{Predicate}\\
       \cmidrule(lr){2-3}\cmidrule(lr){4-5}
       & R@5                 & mR@5                & R@3                              & mR@3                              \\
      \midrule
      CLIP ViT-B/32  & 0.77 & 0.43 & 0.96 & 0.62 \\
      CLIP ViT-L/14  & 0.77 & 0.43 & 0.96 & 0.62 \\
      CLIP ViT-L/14 (PCA)  & 0.76 & 0.41 & 0.96 & 0.61 \\
      BERT \cite{devlin2018bert}  & 0.74 & 0.38 & 0.95 & 0.61 \\
\bottomrule
\end{tabular}
\caption{\textbf{Text model ablation.} We ablate the effects of different text models and their embedding space. We find that CLIP works better than BERT, however different transformer sizes do not affect the final 3D scene graph prediction.}
\label{tab:lang_model}
\vspace{-0.3cm}
\end{table}
}
\newcommand{\figureteaser}{
\begin{figure}
    \vspace*{-0.2cm}
    \centering
    \includegraphics[width=\linewidth]{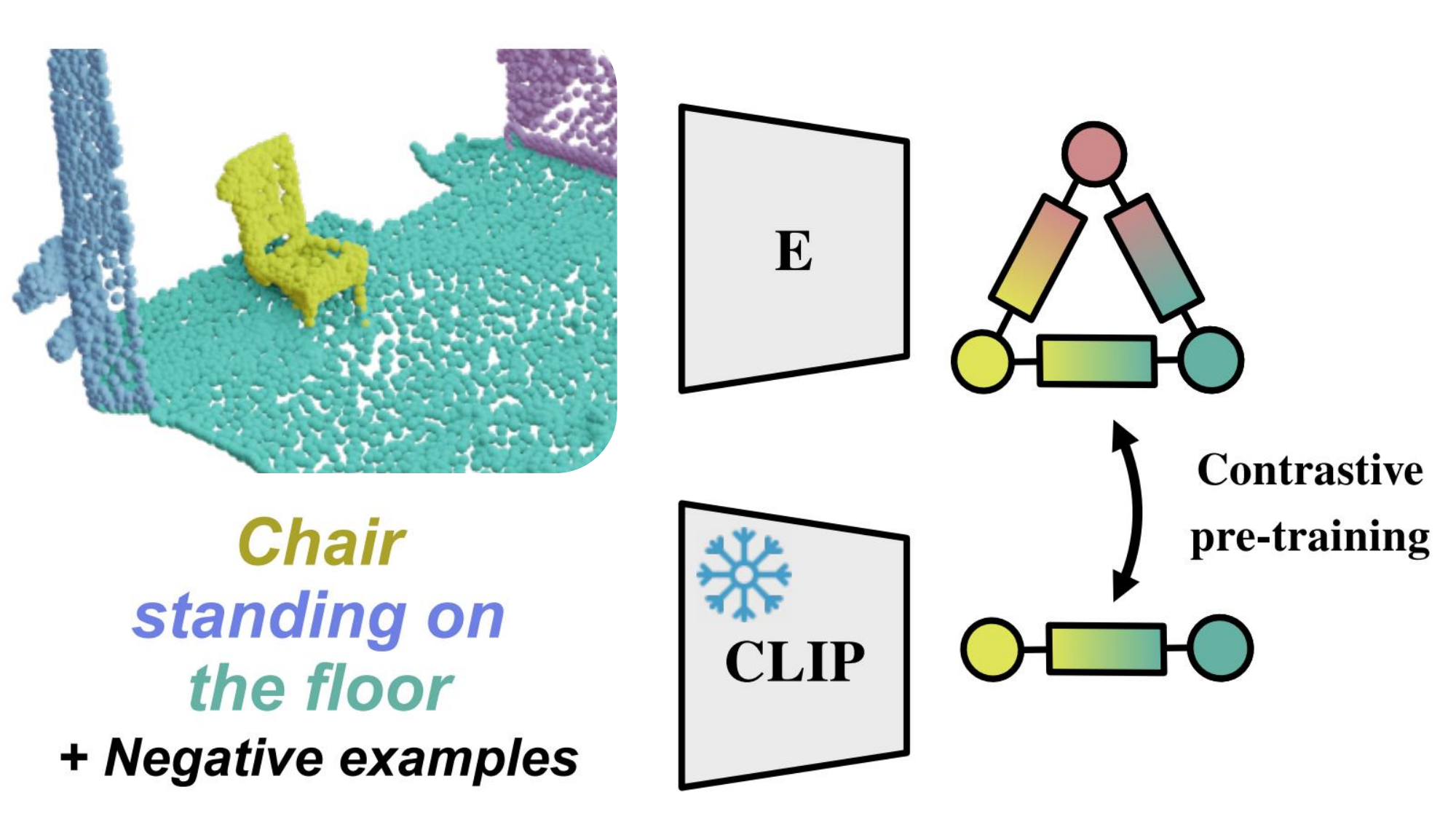}
    \caption{\textbf{\ours key idea.} \ours exploits the natural relatedness of language and 3D scene graphs by pre-training on contrastive language supervision.}
    \label{fig:enter-label}
    \vspace{-0.5cm}
\end{figure}
}

\newcommand{\figuremethod}{
\begin{figure*}
    \centering
    \includegraphics[width=\textwidth]{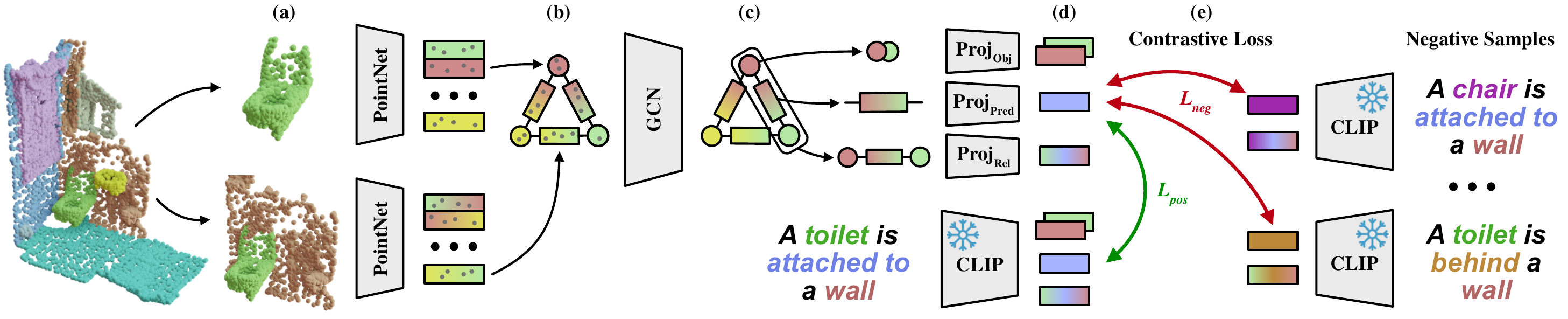}
    \caption{\textbf{Overview of our \ours pre-training framework.} Our method takes as input a class-agnostic segmented point cloud and extracts point sets of objects and pairs of objects (a). The point sets are passed into a PointNet backbone to construct an initial feature graph (b). Using a GCN, the features in the graph get refined (c) and node, edge and node-edge-node triplets are projected into the language feature space (d). Using a contrastive loss, we align the 3D graph features with the CLIP embeddings of the scene description (e).}
    \label{fig:method-fig}
\end{figure*}
}

\newcommand{\figureexamples}{
\begin{figure*}
    \centering
    \includegraphics[width=0.98\textwidth]{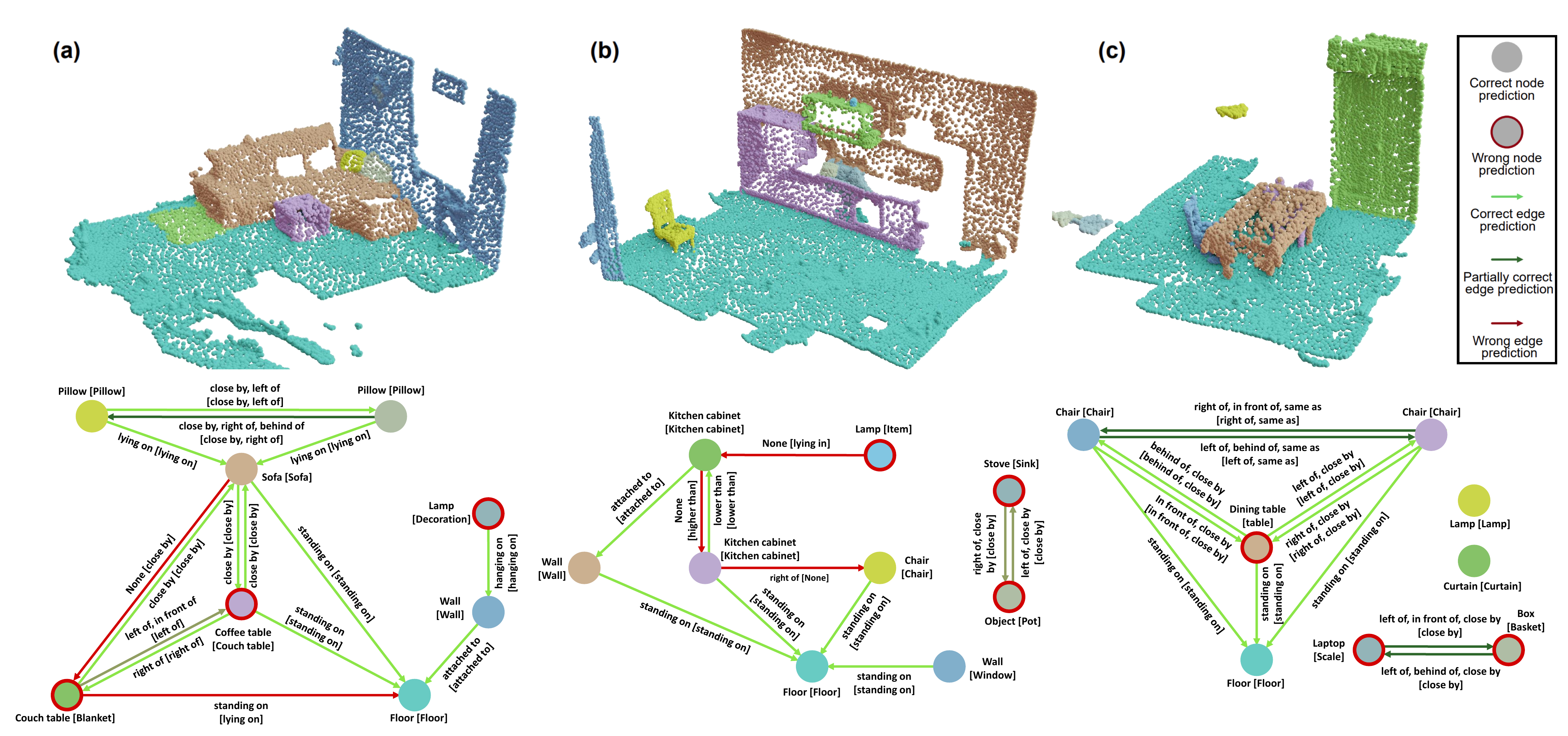}
    \caption{\textbf{3D scene graph visualizations for 3DSSG scene splits.} Qualitative results of 3D scene graph prediction with \ours for three different example scenes. We visualize the top-1 object class prediction for each node and the predicates with a probability greater than 0.5 for each edge. Ground truth labels are shown in square brackets. 
    }
    \label{fig:main_vis}
\end{figure*}
}

\newcommand{\figuretsne}{
\begin{figure*}
    \centering
    \begin{subfigure}{0.245\linewidth}
    \includegraphics[width=0.99\textwidth]{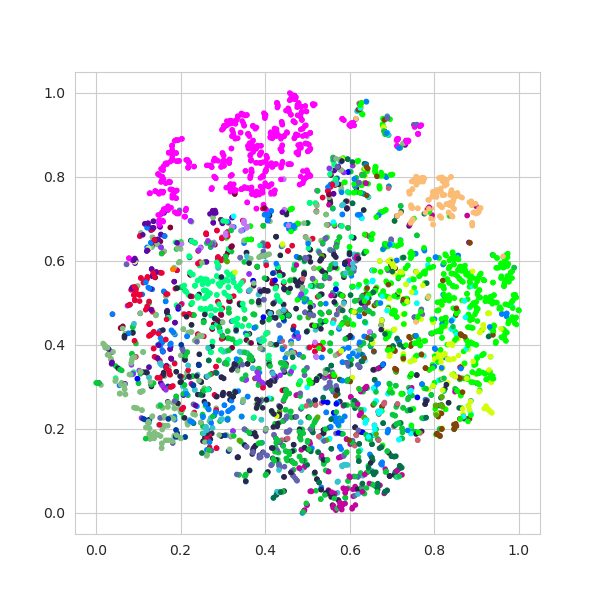}
    \caption{nodes w/o pre-train}
    \end{subfigure}
    \begin{subfigure}{0.245\linewidth}
    \includegraphics[width=0.99\textwidth]{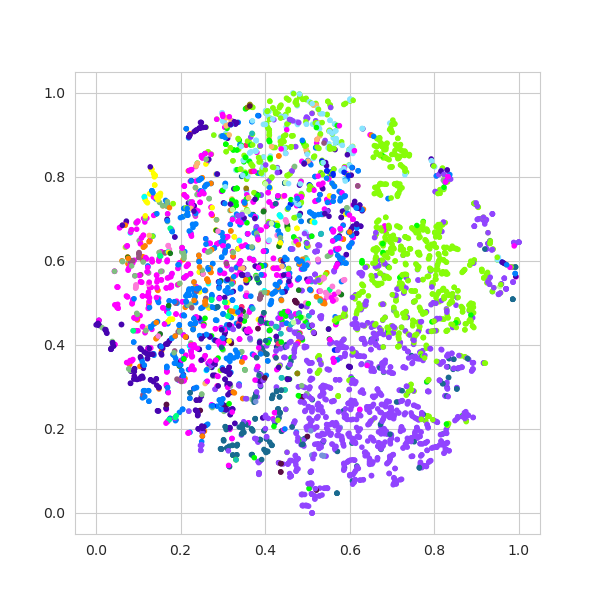}
    \caption{edges w/o pre-train}
    \end{subfigure}
    \begin{subfigure}{0.245\linewidth}
    \includegraphics[width=0.99\textwidth]{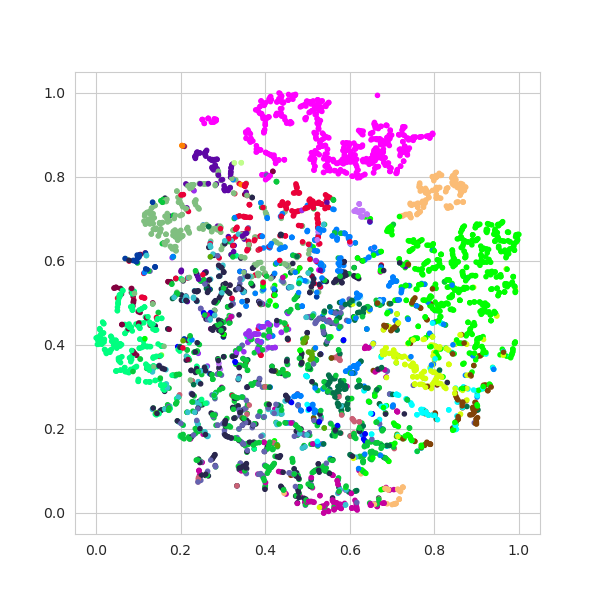}
    \caption{nodes w/ pre-train}
    \end{subfigure}
    \begin{subfigure}{0.245\linewidth}
    \includegraphics[width=0.99\textwidth]{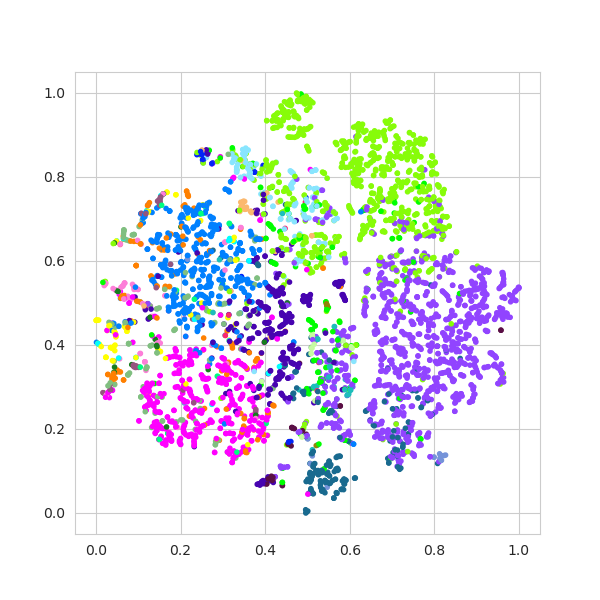}
    \caption{edges w/ pre-train}
    \end{subfigure}
    \caption{\textbf{Learned latent representation.} We show a comparison of the learned representation for supervised training (a)-(b) and our CLIP-based pre-training (c)-(d). Using the language-based contrastive pre-training, our latent representation of objects and predicates is well-structured compared to the model with only supervised training.}
    \label{fig:tsne}
\end{figure*}
}

\newcommand{\figurezeroshot}{
\begin{figure}
    \centering
    {\includegraphics[width=0.8\linewidth]{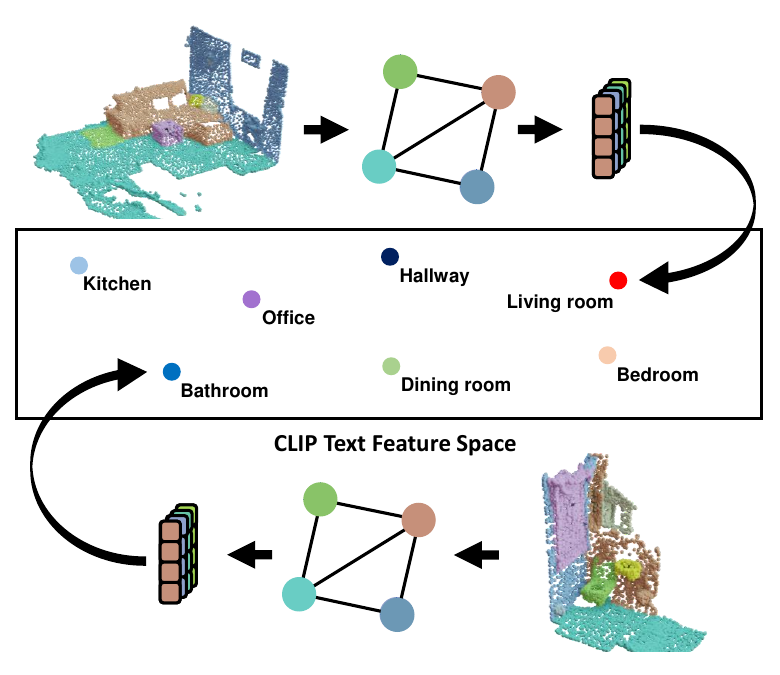}}
    \vspace{-0.2cm}
    \caption{\textbf{Zero-shot room type classification.} Utilizing the language-aligned graph features we can classify the room type of a scene by similarity scoring the feature embedding of a room description and our 3D graph features. Two successful classification examples are shown.}
    \label{fig:room-class}
    \vspace{-0.3cm}
\end{figure}
}

\begin{document}
\maketitle
\vspace*{-0.2cm}
\begin{abstract}
3D scene graphs are an emerging 3D scene representation, that models both the objects present in the scene as well as their relationships. However, learning 3D scene graphs is a challenging task because it requires not only object labels but also relationship annotations, which are very scarce in datasets.
While it is widely accepted that pre-training is an effective approach to improve model performance in low data regimes, in this paper, we find that existing pre-training methods are ill-suited for 3D scene graphs. To solve this issue, we present the first language-based pre-training approach for 3D scene graphs, whereby we exploit the strong relationship between scene graphs and language. To this end, we leverage the language encoder of CLIP, a popular vision-language model, to distill its knowledge into our graph-based network. We formulate a contrastive pre-training, which aligns text embeddings of relationships (subject-predicate-object triplets) and predicted 3D graph features. Our method achieves state-of-the-art results on the main semantic 3D scene graph benchmark by showing improved effectiveness over pre-training baselines and outperforming all the existing fully supervised scene graph prediction methods by a significant margin.
Furthermore, since our scene graph features are language-aligned, it allows us to query the language space of the features in a zero-shot manner. In this paper, we show an example of utilizing this property of the features to predict the room type of a scene without further training.
\end{abstract}
    
\vspace{-0.6cm}
\section{Introduction}
In recent years, 3D scene graphs began to emerge as a new graph-based 3D scene representation that has seen a wide range of applications in computer vision and robotics \cite{Armeni_2019_ICCV,hughes2022hydra,Wald_2020_CVPR,Dhamo_2021_ICCV,sarkar2023sgaligner,looper22vsg,Wu_2021_CVPR,Wu_2023_CVPR}. 
In this field, 3D scene graphs are powerful tools since they allow a compact and straightforward formulation to model both objects in the scene and their semantic relationship. 
In fact, they allow for a more high-level description of a 3D scene, as compared to conventional scene representations, such as 3D object detections or segmentation. 
Due to the encoded high-level information, 3D scene graphs can be used to solve various tasks, such as scene understanding or robot interaction, that conventional 3D models struggle with due to their limited understanding of scene semantics. 
However, predicting 3D scene graphs comes with several challenges, such as noisy and incomplete sensor data, as well as ambiguous object and relationship descriptions. 
Furthermore, while large training data sets are readily available for conventional 3D scene representations, training data for learning 3D scene graphs is much scarcer because relationships are harder to annotate.\looseness=-1
\figureteaser

A frequent approach to deal with the challenges of low data regimes is to facilitate pre-training methods, which allow utilizing the existing data more efficiently \cite{Hou_2021_CVPR, Chen_2022_ECCV}. 
While pre-training is popular in point cloud learning, our pre-training analysis for 3D scene graphs indicates that it is not sufficient for this particular case (see \cref{sec:method}). 
While we could see an improvement in learning object predictions, pre-training did not lead to improved results when predicting object relations. 
We hypothesize that this lack of relationship understanding originates from the poor utilization of context cues in the existing pre-training approaches. Such context information can be well represented and propagated using a graph neural network.
Therefore, one of our key insights is to design a pre-training approach with a graph structure in mind. 

Our second key insight is that scene graphs are inherently related to natural language. In language, a subject, a predicate and an object are the fundamental building blocks of a sentence and in scene graphs, the same triplet forms a relationship represented by two nodes and an edge. Recent advancements in large language models demonstrate their ability to abstract vast semantic knowledge in the embedding space. Aligning this embedding space with different modalities such as vision resulted in a paradigm shift in many scene understanding tasks \cite{radford_2021_PMLR, Peng2023OpenScene, conceptfusion}. Inspired by this progress, we propose to leverage the knowledge of the pre-trained language models for 3D scene graph prediction by formulating a language-based contrastive pre-training. 

Thus, we present the following contributions:
\begin{itemize}
    \item We propose a novel language-based contrastive pre-training for the downstream task of 3D scene graph prediction by exploiting the knowledge of language models.
    \item We show that our pre-training improves 3D scene graph prediction of a simple graph neural network to define a new SOTA by outperforming the existing fully-supervised methods.
    \item We demonstrate further capabilities of our approach, by exploiting the language-aligned scene graph features to predict room types in a zero-shot manner.
\end{itemize}
To the best of our knowledge, we are the first to investigate and propose an approach for 3D scene graph pre-training.

\section{Related Work}
\label{sec:related_work}
\noindent \textbf{3D scene graph prediction.}
A 3D scene graph models a scene as a graph by representing objects in the scene as nodes and relationships between objects as edges connecting two nodes. 
Scene graphs were first proposed in the 2D image domain by Johnson \etal~\cite{Johnson_2015_CVPR} but have been adapted to 3D first by Armeni \etal~\cite{Armeni_2019_ICCV}, 
as a hierarchical structure to connect buildings, rooms and objects. 
Wald \etal~\cite{Wald_2020_CVPR} were the first to introduce a 3D semantic scene graph dataset focused on semantic relationships between objects. 
This dataset is built on top of the large-scale 3D dataset 3RScan~\cite{Wald_2019_ICCV} with over one thousand 3D scans. 
Based on this dataset, some subsequent works focused on expanding the common principles from 2D scene graph prediction to 3D~\cite{Wald_2020_CVPR,Zhang_2021_CVPR}. 
Other works focused on utilizing 3D scene graphs for image and scene retrieval~\cite{Wald_2020_CVPR}, 3D scene reconstruction~\cite{koch2023sgrec3d}, generation and manipulation~\cite{Dhamo_2021_ICCV}, alignment of 3D scene graphs as well as registration of scans~\cite{sarkar2023sgaligner} and change forecasting within the 3D scene~\cite{looper22vsg}. 
Some approaches investigated the construction of 3D scene graphs during dynamic explorations of the scene with \mbox{RGB-D}~\cite{Wu_2021_CVPR} or RGB cameras~\cite{Wu_2023_CVPR}. 
Finally, other works focused on the improvement of the 3D scene graph prediction with advanced message passing and graph convolutions~\cite{Zhang_2021_CVPR}, transformers~\cite{lv2023revisiting} using pre-trained oracle models \cite{wang2023vl}. Instead, our approach focuses on a novel pre-training strategy leveraging the unique similarity between scene graphs and language.

\noindent \textbf{Pre-training for 3D scene understanding.}
In the 2D domain, it is common practice to use backbone networks pre-trained on ImageNet~\cite{Deng_2009_CVPR}, or pre-trained using other representation learning techniques~\cite{NEURIPS2019_ddf35421,Misra_2020_CVPR,pmlr-v119-henaff20a,Wu_2018_CVPR,tian2020contrastive,He_2020_CVPR}. 
Inspired by the progress in the 2D scene understanding, recent works explore to adapt pre-training~\cite{Chen_2021_ICCV,Hassani_2019_ICCV,sanghi2020info3d,sauder2019self,Wang_2021_ICCV} on the 3D object-centric datasets such ShapeNet~\cite{Chang_2015_CORR} and \mbox{ModelNet}~\cite{Wu_2015_CVPR}. %
However, Xie \etal~\cite{Xie_2020_ECCV} showed empirically that pre-training on datasets like ShapeNet~\cite{Chang_2015_CORR} seem to be ineffective for scene-level 3D perception tasks such as 3D segmentation or 3D object detection.
This motivated other works to investigate 3D representation learning based on self-supervised contrastive learning~\cite{Hou_2021_CVPR,Zhang_2021_ICCV,Xie_2020_ECCV, Huang_2021_ICCV,Chen_2022_ECCV}.
However, so far none of these works has considered 3D scene graph prediction as a downstream task for pre-training.

\noindent \textbf{Language-based 3D scene understanding.} \label{sec:language_understanding}
In recent years natural language has become an important part in 2D scene understanding.
The recent advances of large vision-language models such as CLIP~\cite{radford_2021_PMLR}, ALIGN~\cite{jia_2021_PMLR} and follow-up works \cite{li_2021_cvpr_glip,zhang2022glipv2,pmlr-v162-li22n,yuksekgonul2023when} have made a paradigm shift in scene understanding from images enabling open-vocabulary object classification, detection and semantic segmentation~\cite{ghiasi_2022_eccv, li_2022_iclr, gu_2022_iclr, he2023open, liang_2023_cvpr}. This progress ignited interest in distilling the knowledge of 2D vision-language models into 3D representations~\cite{Peng2023OpenScene, kerr2023lerf, ding2023pla, Xue_2023_CVPR, xue2023ulip2}. Rozenberszki \etal~\cite{rozenberszki2022language} demonstrate the usefulness of semantically rich language features by grounding the 3D representation learning with language. They successfully use this idea for language-based pre-training to tackle the long-tail distribution problem in 3D semantic segmentation. Inspired by this we demonstrate how to leverage the vast semantic knowledge from language models in 3D scene graph pre-training. \looseness=-1

\section{3D scene graph pre-training}
\label{sec:method}

\figuremethod

\subsection{Point cloud-based scene graphs pre-training}
Latest improvements in pre-training methods focused on point clouds yield to 
positive results for various applications such as segmentation or object detection~\cite{Hou_2021_CVPR,Zhang_2021_ICCV,Choy_2019_CVPR,Xie_2020_ECCV, Huang_2021_ICCV,Chen_2022_ECCV,rozenberszki2022language}. 
However, the domain of 3D scene graphs has received little attention in pre-training studies. 
To fill this gap, we conduct a pilot study presented in \cref{tab:pretrain_study}, exploring the effectiveness of point cloud pre-training for 3D scene graph prediction.
In this study, we take two recent point cloud-based pre-training methods STRL~\cite{Choy_2019_CVPR} and DepthContrast~\cite{Zhang_2021_ICCV} that open-sourced pre-trained PointNet++~\cite{Qi_2017_NEURIPS} backbones trained on ScanNet \cite{Dai_2017_CVPR} and fine-tune them for scene graph prediction by adding two prediction heads for objects and predicates.  
Additionally, we establish a simple graph-based baseline inspired by the work of Wald \etal~\cite{Wald_2020_CVPR} as a reference. 

For evaluation, we employ top-k recall metrics for objects and predicates, where higher scores indicate better performance. Detailed information regarding the metrics used in this study can be found in \cref{sec:exp_setup}.
Our findings in \cref{tab:pretrain_study} demonstrate that the pre-trained methods perform well on object prediction compared to the graph-based baseline. However, the predicate predictions do not show the same improvement with pre-training compared to the graph-based baseline. 
This indicates that having a graph-based backbone is essential for predicate prediction and that existing pre-training strategies are ineffective for scene graphs since they do not encode graph structures.

This result motivates us to design a pre-training tailored for scene graph prediction with a graph backbone in mind. In the following section, we will introduce our architectural setup as well as our 3D scene graph pre-training approach.

\subsection{Language-based scene graph pre-training}
Since a graph-based backbone seems important for scene graph prediction and the success of pre-training as concluded from our pilot study, we first define our graph-based backbone. 
We start by describing the feature extraction and graph construction methodology to embed a 3D scene into an initial feature graph. 
Then, we continue with a simple graph convolutional network (GCN). 
Furthermore, we specify the modules for contrastive self-supervised pre-training.

\noindent \textbf{Feature extraction and graph construction.}
For the first step of our approach, we build an initial graph $\mathcal{G} = (\mathcal{N},\mathcal{E})$ from a generic scene $s$, 
where $\mathcal{N}$ describes the set of objects and $\mathcal{E}$ describes the relationships of the scene. 
This step involves generating a class-agnostic instance mask $\mathcal{M}$ to extract instances $i$ from the point cloud $\mathcal{P}$ using the mask $\mathcal{M}_i$ from an off-the-shelf instance segmentation method such as Mask3D~\cite{Schult_2022_CORR}. 
Those masks are used to extract from the scene point cloud $\mathcal{P}$ a subset of points $\mathcal{P}_i$ belonging to the object instance $i$, using the relative mask $\mathcal{M}_i$. From $\mathcal{P}_i$  we also produce the object bounding box $\mathcal{B}_i$ and discard any predicted class labels.
Alternatively, when available, ground truth instance annotations can be used to extract the needed instances. 
Each point set $\mathcal{P}_i$ including its color information is fed into a shared PointNet~\cite{Qi_2017_CVPR} to extract features $\phi_n$ for each object node.

To generate edge features $\phi_p$, we use every instance pair $\langle i,j\rangle \in \|M\|\times \|M\|$ to get the combined point set $\mathcal{P}_{ij}$ belonging to the union of their respective bounding boxes $\mathcal{B}_{ij} = \mathcal{B}_i \cup \mathcal{B}_j$. 
Note that, we use the bounding box union to include also points $p_k$ around both objects, which might introduce further contextual information. 
Before feeding every point set $\mathcal{P}_{ij}$ and its color information into a second shared PointNet to extract the edge features $\phi_p$, we concatenate it with a point-wise mask equal to 1 if the point corresponds to object $i$, 2 if the object corresponds to object $j$, and 0 otherwise. \looseness=-1

\tablepilotstudy

\noindent \textbf{Encoder.}
From the extracted features $\phi_n$ and $\phi_p$, we construct an initial feature graph where every node contains only local features describing the object and each edge contains only feature information about a pair of objects. 
However, this information lacks global scene context, which is necessary for predicting complex object relationships. 
To address this issue, we employ a graph convolutional network with message passing to propagate information through the graph such that each node and edge have contextual information about its nearest neighbors. 
To this purpose, we arrange the nodes and edges as triples $t_{ij} = \langle \phi_{n,i},\phi_{p_{ij}},\phi_{n,j} \rangle$. 
Every GCN layer $l_g$ propagates information through the graph in three steps with a message passing procedure similar to~\cite{Wald_2020_CVPR}.
First the triplet $t_{ij}$ is fed into a MLP $g_1(\cdot)$
\begin{equation}
    \left(\psi_{n,i}^{(l_g)},\phi_{p,ij}^{(l_g+1)},\psi_{n,j}^{(l_g)}\right) = g_1\left(\phi_{n,i}^{(l_g)},\phi_{p,ij}^{(l_g)},\phi_{n,j}^{(l_g)}\right)
    \label{eq:gcn_propagate}
\end{equation}
where $\phi_{p,ij}^{(l_g+1)}$ is the updated edge feature and $\psi$ represents the incoming features for the nodes $i$ and $j$. 
Using an aggregation function, the incoming node features are aggregated in a second step. 
We choose the average function as a suitable aggregation function
\begin{equation}\small
    \rho_{n,i}^{(l_g)} = \frac{1}{N_i}\left(\sum_{k\in \mathcal{R}_i}\psi_{n,k}^{(l_g)} + \sum_{k\in \mathcal{R}_j}\psi_{n,k}^{(l_g)}\right)
    \label{eq:gcn_aggregate}
\end{equation}
where $N_i$ denotes the number of edges connected to node $i$, and $\mathcal{R}_i$ and $\mathcal{R}_j$ are the set of nodes connected to node $i$ and node $j$ respectively.

Finally, the aggregated node features $\rho_{n,i}^{(l_g)}$ are passed into a second MLP $g_2(\cdot)$ adding a residual connection:
\begin{equation}\small
    \phi_{n,i}^{(l_g+1)} = \phi_{n,i}^{(l_g)} + g_2\left(\rho_{n,i}^{(l_g)}\right).
    \label{eq:gcn_residual}
\end{equation}
This process is repeated for $k$ layers with which the receptive field of each node grows to finally get the refined features \(\phi_{n,i}^{(l_k)},\phi_{p,ij}^{(l_k)},\phi_{n,j}^{(l_k)}\) containing contextual information of their neighbors and beyond.

\noindent \textbf{Projection heads}
Following Grill \etal~\cite{grill2020bootstrap}, we propose three projector heads that project graph-based features into a high-dimensional feature space that matches the dimensionality $D$ of our text encoder features. 

The three 3-layers MLP project into the projection space the features generated after performing $k$ iterations of graph convolutions.
The first projector $p_1(\cdot)$ projects only the node features, the second $p_2(\cdot)$ only the edge features, while 
we feed the third projector $p_3(\cdot)$ with the concatenated features for each triplet in the graph
to get a singular triplet feature representing the entire relationship.
\begin{equation}
    \footnotesize
    \begin{split}
    &f_{n,i} = p_1\left(\phi_{n,i}^{(l_k)}\right),\,
    f_{p,ij} = p_2\left(\phi_{p,ij}^{(l_k)}\right),\\
    &f_{triplet,ij} = p_3\left(\phi_{n,i}^{(l_k)}\oplus\phi_{p,ij}^{(l_k)}\oplus\phi_{n,j}^{(l_k)}\right).
    \end{split}
\end{equation}
\\
\noindent \textbf{Text Encoder}
We leverage a pre-trained language model to map semantic relationship descriptions to text features. 
For this, we choose CLIP~\cite{radford_2021_PMLR}. CLIP has been shown to have excellent visual understanding capabilities thanks to its vision-language pre-training. 
Note that our approach is agnostic to the choice of language model, but we found CLIP's representations formed by its multi-modal training well-suited to our relationship pre-training. 
During pre-training, we keep the text encoder frozen.

We provide three types of text prompts to our text encoder matching the three projected features from our 3D graph network. 
The first query contains only the object names, the second one provides only the predicate category to the text encoder, and the third query consists of the entire relationship in the form of ``A scene of a [subject] is [predicate] a [object]'' template. 

Each text is then tokenized and encoded to their text embeddings $f_1^t, f_2^t, ..., f_n^t \in \mathbb{R}^D$ where D is the dimensionality of the text representation space. 
Note that we consider scene graphs where a pair of objects can share zero, one or multiple relationships. 
In case no relationship is present, we encode the predicate ``and'' as a neutral predicate to provide a target text embedding for the edge.

\subsection{Language-based pre-training}
\noindent \textbf{Contrastive loss.}
For feature learning, we formulate a contrastive objective between the embeddings of the text model and the predicted 3D graph features by our network. We adopt the cosine similarity as our distance metric. This choice is inspired by CLIP~\cite{radford_2021_PMLR} which has shown that it is a good distance metric for multi-modality contrastive learning as it provides more flexibility compared to $l_1$, $l_2$ or MSE metrics:
\begin{equation}
    \cos(f_i,f_{h(i)}^t) = \dfrac{f_i\cdot f_{h(i)}^t}{|f_i|\cdot |f_{h(i)}^t|}
    \label{eq:cosine_sim}
\end{equation}
where $h(i)$ is the semantic text label for node/edge/triplet $i$. 

During training, we differentiate between positive and negative samples. For the positive samples, our goal is to maximize the cosine similarity between our 3D graph feature encoding and the well-structured feature space of the text model. We do this by minimizing the following term
\begin{equation}
    \mathcal{L}_{pos} = \sum_{i=1}^N \frac{1}{|K|} \sum_{j\in K} 1 - \cos(f_i,f_{h(j)}^t)
\end{equation}
where N is the number of nodes/edges and K is the number of positive samples per node and edge. For objects $|K|=1$ since each node only maps to a singular object, but for edges $|K|\geq 1$ because edges in 3D scene graphs can model more than one predicate/relationship.

Using the negative samples, our goal is to minimize the cosine similarity between our predicted 3D graph feature and the text feature
\begin{equation}
    \mathcal{L}_{neg} = \sum_{i=1}^N \frac{1}{|M|} \sum_{j\in M} \max \left(0, \cos(f_i,f_{h(j)}^t) - \tau \right)
\end{equation}
where $\tau$ is the negative margin and $M \subset C$ consists of negative samples which are a set of labels different from $i$ with $C$ being the set of class label ids in the dataset.

In literature, negative samples are needed to prevent a collapse of the embedding space in most contrastive methods. However, in our case, since we are trying to distill knowledge from the text model it is not strictly necessary, but experimentally we found using negative samples improves the learned representation. 
We provide experimental evidence for improved representation learning using negatives in the supplementary.
Our final language-3D graph pre-training loss is then:
\begin{equation}
    \mathcal{L} = \mathcal{L}_{pos} + \lambda_{neg} \mathcal{L}_{neg}
\end{equation}

\figureexamples

\noindent \textbf{Hard Negative Samples.}
We found that the selection of negative samples is
very important for the quality of the learned representation with our contrastive learning.
Thus, we pick the negative samples from the existing label set on our scene graph dataset.%
Both for objects and predicates we sample a random set of labels from the ones available in our dataset. However, for predicates,
we make sure that the ``and'' predicate, which represents no relation, is always in the negative samples if it is not in the positive keys. By doing so, we enforce the boundary between existing and non-existing predicates. 
For negative relationship samples, we provide hard negatives by taking the true relationship as our source and modifying objects and predicates individually such that a negative relationship sample always shares one object or predicate with the positive sample (see right side of \cref{fig:method-fig}).

\subsection{Scene graph fine-tuning}
After pre-training using the text embeddings, the model needs to be fine-tuned using a supervised loss. This will allow to predict valid object class labels and predicate categories for the generated 3D scene graph. To do so, we discard the projectors and replace them with classification heads. We propose two classification heads, the first to classify the nodes in the graph and the second for predicting predicate labels for each edge. We train the classification heads using a cross-entropy loss $L_{obj}$ for the object nodes and a per-class binary-cross-entropy loss $L_{pred}$ for the predicates:
\begin{equation}
    \mathcal{L} = \lambda_{obj} \mathcal{L}_{\text{obj}} 
    + \lambda_{pred} \mathcal{L}_{\text{pred}}
\end{equation}

\noindent \textbf{Implementation Details.}
During the pre-training, we follow the approach described in the previous section. We choose \textit{CLIP ViT-B/32} with the published weights from OpenAI \cite{radford_2021_PMLR} as our text model rather than larger CLIP models as a good compromise between the text understanding ability and inference speed. The text encoder of the \textit{CLIP ViT-B/32} model provides features of dimensionality $D=512$, which we match with our feature projectors.
During the pre-training stage, we train our model for $50$ epochs until convergence, with the Adam optimizer with a learning rate of 1e-3, linear learning rate decay and a batch size of $6$. We choose the number of negative samples $M=16$, which we randomly sample from $160$ object and $27$ predicate categories. We set the negative loss weight to $\lambda_{neg}=1$ and use a negative margin of $\tau=0.5$.

After pre-training, we proceed with fine-tuning the pre-trained 3D graph backbone using the available 3D scene graph labels. We use the same Adam optimizer with a learning rate of 1e-4, a batch size of $4$ and train for $20$ epochs. To ensure a balanced learning of object and predicate relationships, we set $\lambda_{obj}=0.1$ and $\lambda_{pred}=1.0$,

\section{Experiments}
\label{sec:experiments}
\vspace{-0.1cm}
\subsection{Experimental Setup}
\label{sec:exp_setup}
\noindent \textbf{Dataset.} To validate the effects of our proposed 3D scene graph pre-training, we choose to evaluate our method after fine-tuning on publicly available 3D scene graphs datasets.
The 3DSSG dataset~\cite{Wald_2020_CVPR} is at the time of writing this paper the only large-scale 3D dataset that provides semantic 3D scene graph labels with extensive relationship annotations.
Another 3D scene graph dataset is \cite{Armeni_2019_ICCV}, however, the scene graphs modeled in this dataset focus on hierarchical structuring and lack semantic relationship labels.
In contrast, the most popular 3DSSG dataset provides semantic graph annotations with 160 object classes and 27 relationship categories over more than 1,000 indoor 3D point clouds reconstructions. 
These reconstructions are further subdivided into smaller scene graph splits with up to nine objects per split, resulting in more than 4,000 samples for training and evaluation. 
We follow Wald \etal~\cite{Wald_2019_ICCV} and use the original train/validation splits for training and evaluation.
At the time of writing this paper, 3DSSG is the only large-scale 3D dataset that provides semantic 3D scene graph labels with extensive relationship annotations.

\noindent \textbf{Evaluation metrics.}
For comparison with existing methods that predict 3D scene graphs without pre-training, we follow \cite{wald2022learning,Zhang_2021_CVPR,Wald_2020_CVPR,Xu_2017_CVPR,Yang_2018_ECCV} and evaluate object and predicate predictions separately. Additionally, we jointly evaluate subject-predicate-object triples as relationships formed from two nodes in the graph and their enclosing edge. 
Since our approach predicts objects and predicates independently, we follow Yang \etal~\cite{Yang_2018_ECCV} and multiply the predicted object node and predicate edge probabilities to obtain a scored list of triplet predictions.
We rank the triples by their score and use a top-k recall metric (R@k) \cite{Lu_2016_ECCV} for our main evaluation with existing works. 
The top-k recall metrics are used since one edge in the scene graph can represent multiple ground truth relationships. 
In our ablations, we additionally provide results on the less used top-k mean recall metric \cite{Chen_2019_CVPR} for objects and predicates
to cope with the high class imbalance present which affects the dataset, as shown in~\cite{Wald_2020_CVPR}. 
Using this metric gives better performance indication for underrepresented classes in the dataset.

\vspace{-0.14cm}
\subsection{3D scene graph prediction}
\vspace{-0.14cm}
In \cref{tab:main_evaluation} we report the results of our pre-trained model as described in \cref{sec:method}. We compare against recent non-pretrained 3D scene graph methods (SGGPoint \cite{Zhang_2021_CVPR}, 3DSSG \cite{Wald_2020_CVPR}, SGFN \cite{Wu_2021_CVPR}, Liu et al. \cite{Liu_2022_TVCG}) and adapted 2D scene graph methods (MSDN \cite{Li_2017_ICCV}, KERN \cite{Chen_2019_CVPR}, BGNN \cite{Li_2021_CVPR}). For the 2D scene graph methods, the 2D object detector was replaced by a PointNet-based feature extractor. 
\cref{tab:main_evaluation} shows that we outperform all existing methods on the most used 3D scene graph prediction metrics. The exception is SGFN~\cite{Wu_2021_CVPR}, which reports equal results in predicate predictions for R@5 and slightly better results for R@3. But especially for object classification, we achieve to outperform all other methods by a considerable margin with +7\%/+4\% improvements to our closest competitor SGFN~\cite{Wu_2021_CVPR}. For relationship prediction overall, we also outperform all other scene graph methods by a large margin for most methods and a considerable margin for SGFN. Regarding the close performance to SGFN, we hypothesize that for the predicates, we reached a saturation point on this dataset with these metrics. This is especially apparent for the R@5 metric, where we and SGFN both score 99\%. 
In the following ablation, we try to overcome this saturation drawback by adopting the mR@k metric which has been less frequently used in literature.
In \cref{fig:main_vis}, we provide qualitative examples of our 3D scene graph predictions for a diverse set of 3D scenes.
We show the top-1 prediction for nodes and predicates that have a prediction score over 0.5.
Our method is able to predict 3D scene graphs with a high accuracy.
Nodes are generally classified correctly, but for a few incorrectly classified objects, a label that nevertheless fits the context of the scene is chosen.
Additionally, we observe that small objects with only a few relationships to others are more often misclassified, indicating that the graph nature is beneficial for object classification.
Predicates are also predicted with a high accuracy, still, we observe that incorrectly predicted edges often coincide with misclassified nodes which propagate incorrect information. 
\tablemainevaluation
\tablepretrainevaluation
\figuretsne
\vspace{-0.10cm}
\subsection{Ablations}
\vspace{-0.15cm}
\noindent \textbf{Pre-training comparison.} In \cref{tab:pretrain_evaluation}, we compare our pre-training with selected recent point cloud pre-training methods and a graph baseline that has the same graph backbone as our method but is not pre-trained using our approach. 
First, we want to highlight the large improvement between our method with pre-training and the graph baseline. 
Using our pre-training we are able to improve object classification by +14\% for R@5 and +13\% for the mR@5 metric. 
Predicate prediction improves similarly by +2\% on the R@3 metric and +10\% on the mR@3 metric.
This indicates that our pre-training is especially effective for rare predicates. 
The improvements compared to point cloud-based pre-trainings are also large. 
While point cloud-based pre-trainings were able to improve object classification, but failed to improve predicate predictions for 3D scene graph predictions, our pre-training is effective for both tasks, outperforming STRL \cite{Choy_2019_CVPR} and DepthContrast \cite{Zhang_2021_ICCV} with a considerable margin of +7\%/+8\% on mR@5 object classification and +16\%/+17\% on mR@3 predicate prediction.
We provide an additional ablation to our method with no graph backbone.
This method demonstrates better performance than the non-pre-trained graph baseline, indicating the effectiveness of our pre-training. 
But compared to our method with graph-backbone, the model without a graph-backbone performs much worse, especially on predicate prediction, confirming our second takeaway from our pilot study that a graph-backbone is essential for scene graph prediction.
\tablepretrainsupervision

\noindent \textbf{Learned Representation. }
In \cref{fig:tsne}, we analyze the pre-trained representation space for objects and predicates by visualizing a t-SNE projection of the learned features. As a comparison, we additionally provide feature projections for our method fine-tuned on 3DSSG without pre-training. Using our pre-training, we have learned a more structured 3D feature representation for objects and predicates, by having anchored our 3D graph features to the well-structured text embedding. Having this structured latent representation allows us to achieve significant improvements in the downstream task of 3D scene graph prediction when fine-tuning from this embedding space.

\noindent \textbf{Text supervision. }
During pre-training, we provide three types of text embeddings as our supervision signal.
First, the text embedding of objects, second the text embedding of predicates and third a composed embedding of the relationship in the form ``A scene of a [subject] is [predicate] a [object]".
In \cref{tab:pretrain_supervision}, we ablate the effects for each target text embedding. 
Providing the composed relationship form already contains all the information about the objects and predicates, but we observe that only providing the relationship as supervision produces inferior results compared to providing individual embeddings for objects and predicates.
We assume that this corresponds to the issue that CLIP and related models are not good at understanding complex compositional scenes~\cite{yuksekgonul2023when,Ma_2023_CVPR,Doveh_2023_CVPR}. 
In our supplementary, we provide further analyses to examine which parts in relationships text descriptions CLIP attends to. 
Combining object, predicate, and relationship embeddings during pre-training results in the best pre-trained model.

\noindent \textbf{Effect of the language model.}  In \cref{tab:lang_model}, we consider alternative language models to our selected \textit{CLIP ViT-B/32} model. 
We consider BERT \cite{devlin2018bert}, which is a popular language model trained on large amounts of text data, rather than multi-modal image-text training from CLIP with its variant \textit{bert\_uncased\_L-8\_H-512\_A-8} from \cite{turc2019well}. 
We also choose \textit{CLIP ViT-L/14} which is another CLIP model, but with a larger text transformer. 
The language features from \textit{CLIP ViT-L/14} have a dimensionality of $D=768$. 
We, therefore, have to adapt the architecture of our projectors to match the dimensionality of the language features from \textit{CLIP ViT-L/14}. 
Additionally, we try to project the 768-dimensional features to 512 dimensions using PCA. 
We find CLIP's rich embedding structure from the multi-model training produces better results compared to the text-based-only model BERT. 
However, we observe that using a larger CLIP model does not improve the pre-training effectiveness. 
The projected CLIP features using PCA produce slightly worse results. 
We assume this is because some information in the CLIP embedding gets lost when projecting its latent space to a lower dimension.

\tableablationcbertclip

\subsection{Zero-shot room classification}
In \cref{fig:room-class} we show one of many possible use cases granted by our language-aligned scene graph features. 
Given the unique property of our 3D scene graph method representing node and edge features in the well-structured CLIP representation space, we are able to query the graph in a zero-shot manner. 
One use case that leverages this representation, is querying the room type of the scene. 
Here, we exploit the fact that the features in our 3D scene graph are aligned with the language features from objects that have a high feature similarity with the room that they correspond to. 
For this, we encode candidate room types like \textit{kitchen}, \textit{bathroom}, \textit{living} room, \etc using the CLIP language encoder to get the features $f_1^q, f_2^q, ..., f_n^q \in \mathbb{R}^D$ for each room query. 
Then we encode a scene $s$ using our language-aligned graph backbone and use average-polling to pool the features from all nodes in the graph $f_{SG} = \varphi(f_{n,1},f_{n,2},...,f_{n,k})$, where $\varphi : \mathbb{R}^{K\times D \rightarrow D}$. Finally, we compute the similarity cosine score between the pooled graph feature and the candidate room types and select the room type with the highest similarity score using $\text{argmax}_q\{\cos(f_{SG},f_n^q)\}$. 
\cref{fig:room-class} shows qualitative results for two diverse examples for our room type prediction together with the abstracted methodology. We are able to successfully classify a bathroom and a living room by their language-aligned 3D graph features. Since this experiment is performed in a zero-shot manner, no quantitative results are available, however, we provide more predictions in our supplementary.
\figurezeroshot
This is one of many possible use cases made possible by our language-based pre-training and language-aligned latent graph features. 
Note, that this approach is different from recent open-vocabulary 3D understanding methods such as \cite{Peng2023OpenScene,Xue_2023_CVPR}. 
Although we are able to exploit the relatedness of features in language space, we are not able to query unseen object and predicate classes since our language pre-training was done with a fixed vocabulary. We leave open-vocabulary 3D scene graph prediction to be investigated in future works.

\section{Conclusion}
\label{sec:conclusion}
In this paper, we find that recent pre-training approaches for 3D scene understanding on point clouds are ineffective for 3D scene graph prediction due to the inability to properly represent relations among objects. 
To this end, we introduce \ours, the first graph-based pre-training approach designed explicitly for 3D scene graphs, that exploits the tight connection between scene graphs and natural language. 
In the experimental study, we demonstrate that our method is more effective than existing pre-training baselines and
achieves better performance than SOTA fully-supervised approaches on the 3DSSG dataset. 
Additionally, we show a zero-shot room type prediction use case based on exploiting language-aligned 3D graph features. 
In conclusion, our work contributes to 3D scene graph prediction, which is an important prerequisite for a wide range of downstream applications relying on accurate scene representations.\looseness=-1

\vspace{0.5em}\noindent\textbf{Acknowledgement}
 This work was partly supported by the EU Horizon 2020 research and innovation program under grant agreement No. 101017274 (DARKO).

{
    \small
    \bibliographystyle{ieeenat_fullname}
    \bibliography{main}

\begin{thebibliography}{69}
\providecommand{\natexlab}[1]{#1}
\providecommand{\url}[1]{\texttt{#1}}
\expandafter\ifx\csname urlstyle\endcsname\relax
  \providecommand{\doi}[1]{doi: #1}\else
  \providecommand{\doi}{doi: \begingroup \urlstyle{rm}\Url}\fi

\bibitem[Armeni et~al.(2019)Armeni, He, Gwak, Zamir, Fischer, Malik, and
  Savarese]{Armeni_2019_ICCV}
Iro Armeni, Zhi-Yang He, JunYoung Gwak, Amir~R. Zamir, Martin Fischer, Jitendra
  Malik, and Silvio Savarese.
\newblock 3d scene graph: A structure for unified semantics, 3d space, and
  camera.
\newblock In \emph{Proceedings of the IEEE/CVF International Conference on
  Computer Vision (ICCV)}, 2019.

\bibitem[Bachman et~al.(2019)Bachman, Hjelm, and
  Buchwalter]{NEURIPS2019_ddf35421}
Philip Bachman, R~Devon Hjelm, and William Buchwalter.
\newblock Learning representations by maximizing mutual information across
  views.
\newblock In \emph{Advances in Neural Information Processing Systems}. Curran
  Associates, Inc., 2019.

\bibitem[Chang et~al.(2015)Chang, Funkhouser, Guibas, Hanrahan, Huang, Li,
  Savarese, Savva, Song, Su, Xiao, Yi, and Yu]{Chang_2015_CORR}
Angel~X. Chang, Thomas~A. Funkhouser, Leonidas~J. Guibas, Pat Hanrahan,
  Qi{-}Xing Huang, Zimo Li, Silvio Savarese, Manolis Savva, Shuran Song, Hao
  Su, Jianxiong Xiao, Li Yi, and Fisher Yu.
\newblock Shapenet: An information-rich 3d model repository.
\newblock \emph{CoRR}, abs/1512.03012, 2015.

\bibitem[Chen et~al.(2019)Chen, Yu, Chen, and Lin]{Chen_2019_CVPR}
Tianshui Chen, Weihao Yu, Riquan Chen, and Liang Lin.
\newblock Knowledge-embedded routing network for scene graph generation.
\newblock In \emph{Proceedings of the IEEE/CVF Conference on Computer Vision
  and Pattern Recognition (CVPR)}, 2019.

\bibitem[Chen et~al.(2021)Chen, Liu, Ni, Wang, Yang, Liu, Li, and
  Tian]{Chen_2021_ICCV}
Ye Chen, Jinxian Liu, Bingbing Ni, Hang Wang, Jiancheng Yang, Ning Liu, Teng
  Li, and Qi Tian.
\newblock Shape self-correction for unsupervised point cloud understanding.
\newblock In \emph{Proceedings of the IEEE/CVF International Conference on
  Computer Vision (ICCV)}, pages 8382--8391, 2021.

\bibitem[Chen et~al.(2022)Chen, Nie{\ss}ner, and Dai]{Chen_2022_ECCV}
Yujin Chen, Matthias Nie{\ss}ner, and Angela Dai.
\newblock 4dcontrast: Contrastive learning with dynamic correspondences
  for 3d scene understanding.
\newblock In \emph{Computer Vision -- ECCV 2022}, pages 543--560, Cham, 2022.
  Springer Nature Switzerland.

\bibitem[Choy et~al.(2019)Choy, Gwak, and Savarese]{Choy_2019_CVPR}
Christopher Choy, JunYoung Gwak, and Silvio Savarese.
\newblock 4d spatio-temporal convnets: Minkowski convolutional neural networks.
\newblock In \emph{Proceedings of the IEEE/CVF Conference on Computer Vision
  and Pattern Recognition (CVPR)}, 2019.

\bibitem[Dai et~al.(2017)Dai, Chang, Savva, Halber, Funkhouser, and
  Nie{\ss}ner]{Dai_2017_CVPR}
Angela Dai, Angel~X Chang, Manolis Savva, Maciej Halber, Thomas Funkhouser, and
  Matthias Nie{\ss}ner.
\newblock Scannet: Richly-annotated 3d reconstructions of indoor scenes.
\newblock In \emph{Proceedings of the IEEE/CVF Conference on Computer Vision
  and Pattern Recognition (CVPR)}, pages 5828--5839, 2017.

\bibitem[Deng et~al.(2009)Deng, Dong, Socher, Li, Li, and
  Fei-Fei]{Deng_2009_CVPR}
Jia Deng, Wei Dong, Richard Socher, Li-Jia Li, Kai Li, and Li Fei-Fei.
\newblock Imagenet: A large-scale hierarchical image database.
\newblock In \emph{2009 IEEE Conference on Computer Vision and Pattern
  Recognition}, pages 248--255, 2009.

\bibitem[Devlin et~al.(2018)Devlin, Chang, Lee, and Toutanova]{devlin2018bert}
Jacob Devlin, Ming-Wei Chang, Kenton Lee, and Kristina Toutanova.
\newblock Bert: Pre-training of deep bidirectional transformers for language
  understanding.
\newblock \emph{arXiv preprint arXiv:1810.04805}, 2018.

\bibitem[Dhamo et~al.(2021)Dhamo, Manhardt, Navab, and
  Tombari]{Dhamo_2021_ICCV}
Helisa Dhamo, Fabian Manhardt, Nassir Navab, and Federico Tombari.
\newblock Graph-to-3d: End-to-end generation and manipulation of 3d scenes
  using scene graphs.
\newblock In \emph{Proceedings of the IEEE/CVF International Conference on
  Computer Vision (ICCV)}, pages 16352--16361, 2021.

\bibitem[Ding et~al.(2023)Ding, Yang, Xue, Zhang, Bai, and Qi]{ding2023pla}
Runyu Ding, Jihan Yang, Chuhui Xue, Wenqing Zhang, Song Bai, and Xiaojuan Qi.
\newblock Pla: Language-driven open-vocabulary 3d scene understanding.
\newblock In \emph{Proceedings of the IEEE/CVF Conference on Computer Vision
  and Pattern Recognition}, pages 7010--7019, 2023.

\bibitem[Doveh et~al.(2023)Doveh, Arbelle, Harary, Schwartz, Herzig, Giryes,
  Feris, Panda, Ullman, and Karlinsky]{Doveh_2023_CVPR}
Sivan Doveh, Assaf Arbelle, Sivan Harary, Eli Schwartz, Roei Herzig, Raja
  Giryes, Rogerio Feris, Rameswar Panda, Shimon Ullman, and Leonid Karlinsky.
\newblock Teaching structured vision \& language concepts to vision \& language
  models.
\newblock In \emph{Proceedings of the IEEE/CVF Conference on Computer Vision
  and Pattern Recognition (CVPR)}, pages 2657--2668, 2023.

\bibitem[Ghiasi et~al.(2022)Ghiasi, Gu, Cui, and Lin]{ghiasi_2022_eccv}
Golnaz Ghiasi, Xiuye Gu, Yin Cui, and Tsung-Yi Lin.
\newblock Scaling open-vocabulary image segmentation with image-level labels.
\newblock In \emph{European Conference on Computer Vision}, pages 540--557.
  Springer, 2022.

\bibitem[Grill et~al.(2020)Grill, Strub, Altch{\'e}, Tallec, Richemond,
  Buchatskaya, Doersch, Avila~Pires, Guo, Gheshlaghi~Azar,
  et~al.]{grill2020bootstrap}
Jean-Bastien Grill, Florian Strub, Florent Altch{\'e}, Corentin Tallec, Pierre
  Richemond, Elena Buchatskaya, Carl Doersch, Bernardo Avila~Pires, Zhaohan
  Guo, Mohammad Gheshlaghi~Azar, et~al.
\newblock Bootstrap your own latent-a new approach to self-supervised learning.
\newblock \emph{Advances in neural information processing systems},
  33:\penalty0 21271--21284, 2020.

\bibitem[Gu et~al.(2022)Gu, Lin, Kuo, and Cui]{gu_2022_iclr}
Xiuye Gu, Tsung-Yi Lin, Weicheng Kuo, and Yin Cui.
\newblock Open-vocabulary object detection via vision and language knowledge
  distillation.
\newblock In \emph{International Conference on Learning Representations}, 2022.

\bibitem[Hassani and Haley(2019)]{Hassani_2019_ICCV}
Kaveh Hassani and Mike Haley.
\newblock Unsupervised multi-task feature learning on point clouds.
\newblock In \emph{Proceedings of the IEEE/CVF International Conference on
  Computer Vision (ICCV)}, 2019.

\bibitem[He et~al.(2020)He, Fan, Wu, Xie, and Girshick]{He_2020_CVPR}
Kaiming He, Haoqi Fan, Yuxin Wu, Saining Xie, and Ross Girshick.
\newblock Momentum contrast for unsupervised visual representation learning.
\newblock In \emph{Proceedings of the IEEE/CVF Conference on Computer Vision
  and Pattern Recognition (CVPR)}, 2020.

\bibitem[He et~al.(2023)He, Guo, Dai, Qiao, Shu, Ren, and Xia]{he2023open}
Sunan He, Taian Guo, Tao Dai, Ruizhi Qiao, Xiujun Shu, Bo Ren, and Shu-Tao Xia.
\newblock Open-vocabulary multi-label classification via multi-modal knowledge
  transfer.
\newblock In \emph{Proceedings of the AAAI Conference on Artificial
  Intelligence}, pages 808--816, 2023.

\bibitem[Henaff(2020)]{pmlr-v119-henaff20a}
Olivier Henaff.
\newblock Data-efficient image recognition with contrastive predictive coding.
\newblock In \emph{Proceedings of the 37th International Conference on Machine
  Learning}, pages 4182--4192. PMLR, 2020.

\bibitem[Hou et~al.(2021)Hou, Graham, Niessner, and Xie]{Hou_2021_CVPR}
Ji Hou, Benjamin Graham, Matthias Niessner, and Saining Xie.
\newblock Exploring data-efficient 3d scene understanding with contrastive
  scene contexts.
\newblock In \emph{Proceedings of the IEEE/CVF Conference on Computer Vision
  and Pattern Recognition (CVPR)}, pages 15587--15597, 2021.

\bibitem[Huang et~al.(2021)Huang, Xie, Zhu, and Zhu]{Huang_2021_ICCV}
Siyuan Huang, Yichen Xie, Song-Chun Zhu, and Yixin Zhu.
\newblock Spatio-temporal self-supervised representation learning for 3d point
  clouds.
\newblock In \emph{Proceedings of the IEEE/CVF International Conference on
  Computer Vision (ICCV)}, pages 6535--6545, 2021.

\bibitem[Hughes et~al.(2022)Hughes, Chang, and Carlone]{hughes2022hydra}
N. Hughes, Y. Chang, and L. Carlone.
\newblock Hydra: A real-time spatial perception system for {3D} scene graph
  construction and optimization.
\newblock 2022.

\bibitem[Jatavallabhula et~al.(2023)Jatavallabhula, Kuwajerwala, Gu, Omama,
  Chen, Li, Iyer, Saryazdi, Keetha, Tewari, Tenenbaum, {de Melo}, Krishna,
  Paull, Shkurti, and Torralba]{conceptfusion}
{Krishna Murthy} Jatavallabhula, Alihusein Kuwajerwala, Qiao Gu, Mohd Omama,
  Tao Chen, Shuang Li, Ganesh Iyer, Soroush Saryazdi, Nikhil Keetha, Ayush
  Tewari, {Joshua B.} Tenenbaum, {Celso Miguel} {de Melo}, Madhava Krishna,
  Liam Paull, Florian Shkurti, and Antonio Torralba.
\newblock Conceptfusion: Open-set multimodal 3d mapping.
\newblock \emph{arXiv}, 2023.

\bibitem[Jia et~al.(2021)Jia, Yang, Xia, Chen, Parekh, Pham, Le, Sung, Li, and
  Duerig]{jia_2021_PMLR}
Chao Jia, Yinfei Yang, Ye Xia, Yi-Ting Chen, Zarana Parekh, Hieu Pham, Quoc Le,
  Yun-Hsuan Sung, Zhen Li, and Tom Duerig.
\newblock Scaling up visual and vision-language representation learning with
  noisy text supervision.
\newblock In \emph{International conference on machine learning}, pages
  4904--4916. PMLR, 2021.

\bibitem[Johnson et~al.(2015)Johnson, Krishna, Stark, Li, Shamma, Bernstein,
  and Fei-Fei]{Johnson_2015_CVPR}
Justin Johnson, Ranjay Krishna, Michael Stark, Li-Jia Li, David Shamma, Michael
  Bernstein, and Li Fei-Fei.
\newblock Image retrieval using scene graphs.
\newblock In \emph{Proceedings of the IEEE Conference on Computer Vision and
  Pattern Recognition (CVPR)}, 2015.

\bibitem[Kerr et~al.(2023)Kerr, Kim, Goldberg, Kanazawa, and
  Tancik]{kerr2023lerf}
Justin Kerr, Chung~Min Kim, Ken Goldberg, Angjoo Kanazawa, and Matthew Tancik.
\newblock Lerf: Language embedded radiance fields.
\newblock \emph{arXiv preprint arXiv:2303.09553}, 2023.

\bibitem[Koch et~al.(2023)Koch, Hermosilla, Vaskevicius, Colosi, and
  Ropinski]{koch2023sgrec3d}
Sebastian Koch, Pedro Hermosilla, Narunas Vaskevicius, Mirco Colosi, and Timo
  Ropinski.
\newblock Sgrec3d: Self-supervised 3d scene graph learning via object-level
  scene reconstruction.
\newblock \emph{arXiv preprint arXiv:2309.15702}, 2023.

\bibitem[Li et~al.(2022{\natexlab{a}})Li, Weinberger, Belongie, Koltun, and
  Ranftl]{li_2022_iclr}
Boyi Li, Kilian~Q Weinberger, Serge Belongie, Vladlen Koltun, and Rene Ranftl.
\newblock Language-driven semantic segmentation.
\newblock In \emph{International Conference on Learning Representations},
  2022{\natexlab{a}}.

\bibitem[Li et~al.(2022{\natexlab{b}})Li, Li, Xiong, and Hoi]{pmlr-v162-li22n}
Junnan Li, Dongxu Li, Caiming Xiong, and Steven Hoi.
\newblock {BLIP}: Bootstrapping language-image pre-training for unified
  vision-language understanding and generation.
\newblock In \emph{Proceedings of the 39th International Conference on Machine
  Learning}, pages 12888--12900. PMLR, 2022{\natexlab{b}}.

\bibitem[Li* et~al.(2022)Li*, Zhang*, Zhang*, Yang, Li, Zhong, Wang, Yuan,
  Zhang, Hwang, Chang, and Gao]{li_2021_cvpr_glip}
Liunian~Harold Li*, Pengchuan Zhang*, Haotian Zhang*, Jianwei Yang, Chunyuan
  Li, Yiwu Zhong, Lijuan Wang, Lu Yuan, Lei Zhang, Jenq-Neng Hwang, Kai-Wei
  Chang, and Jianfeng Gao.
\newblock Grounded language-image pre-training.
\newblock In \emph{CVPR}, 2022.

\bibitem[Li et~al.(2021)Li, Zhang, Wan, and He]{Li_2021_CVPR}
Rongjie Li, Songyang Zhang, Bo Wan, and Xuming He.
\newblock Bipartite graph network with adaptive message passing for unbiased
  scene graph generation.
\newblock In \emph{Proceedings of the IEEE/CVF Conference on Computer Vision
  and Pattern Recognition (CVPR)}, pages 11109--11119, 2021.

\bibitem[Li et~al.(2017)Li, Ouyang, Zhou, Wang, and Wang]{Li_2017_ICCV}
Yikang Li, Wanli Ouyang, Bolei Zhou, Kun Wang, and Xiaogang Wang.
\newblock Scene graph generation from objects, phrases and region captions.
\newblock In \emph{Proceedings of the IEEE International Conference on Computer
  Vision (ICCV)}, 2017.

\bibitem[Liang et~al.(2023)Liang, Wu, Dai, Li, Zhao, Zhang, Zhang, Vajda, and
  Marculescu]{liang_2023_cvpr}
Feng Liang, Bichen Wu, Xiaoliang Dai, Kunpeng Li, Yinan Zhao, Hang Zhang,
  Peizhao Zhang, Peter Vajda, and Diana Marculescu.
\newblock Open-vocabulary semantic segmentation with mask-adapted clip.
\newblock In \emph{Proceedings of the IEEE/CVF Conference on Computer Vision
  and Pattern Recognition}, pages 7061--7070, 2023.

\bibitem[Liu et~al.(2022)Liu, Long, Zhang, Liu, Zhang, Yin, and
  Yang]{Liu_2022_TVCG}
Yuanyuan Liu, Chengjiang Long, Zhaoxuan Zhang, Bokai Liu, Qiang Zhang, Baocai
  Yin, and Xin Yang.
\newblock Explore contextual information for 3d scene graph generation.
\newblock \emph{IEEE Transactions on Visualization and Computer Graphics},
  pages 1--13, 2022.

\bibitem[Looper et~al.(2023)Looper, Rodriguez-Puigvert, Siegwart, Cadena, and
  Schmid]{looper22vsg}
Samuel Looper, Javier Rodriguez-Puigvert, Roland Siegwart, Cesar Cadena, and
  Lukas Schmid.
\newblock 3d vsg: Long-term semantic scene change prediction through 3d
  variable scene graphs.
\newblock IEEE International Conference on Robotics and Automation (ICRA),
  2023.

\bibitem[Lu et~al.(2016)Lu, Krishna, Bernstein, and Fei-Fei]{Lu_2016_ECCV}
Cewu Lu, Ranjay Krishna, Michael Bernstein, and Li Fei-Fei.
\newblock Visual relationship detection with language priors.
\newblock In \emph{Computer Vision -- ECCV 2016}, pages 852--869, Cham, 2016.
  Springer International Publishing.

\bibitem[Lv et~al.(2023)Lv, Qi, Li, Yang, and Ma]{lv2023revisiting}
Changsheng Lv, Mengshi Qi, Xia Li, Zhengyuan Yang, and Huadong Ma.
\newblock Revisiting transformer for point cloud-based 3d scene graph
  generation.
\newblock \emph{arXiv preprint arXiv:2303.11048}, 2023.

\bibitem[Ma et~al.(2023)Ma, Hong, Gul, Gandhi, Gao, and Krishna]{Ma_2023_CVPR}
Zixian Ma, Jerry Hong, Mustafa~Omer Gul, Mona Gandhi, Irena Gao, and Ranjay
  Krishna.
\newblock Crepe: Can vision-language foundation models reason compositionally?
\newblock In \emph{Proceedings of the IEEE/CVF Conference on Computer Vision
  and Pattern Recognition (CVPR)}, pages 10910--10921, 2023.

\bibitem[Misra and Maaten(2020)]{Misra_2020_CVPR}
Ishan Misra and Laurens van~der Maaten.
\newblock Self-supervised learning of pretext-invariant representations.
\newblock In \emph{Proceedings of the IEEE/CVF Conference on Computer Vision
  and Pattern Recognition (CVPR)}, 2020.

\bibitem[Peng et~al.(2023)Peng, Genova, Jiang, Tagliasacchi, Pollefeys, and
  Funkhouser]{Peng2023OpenScene}
Songyou Peng, Kyle Genova, Chiyu~"Max" Jiang, Andrea Tagliasacchi, Marc
  Pollefeys, and Thomas Funkhouser.
\newblock Openscene: 3d scene understanding with open vocabularies.
\newblock In \emph{Proceedings of the IEEE/CVF Conference on Computer Vision
  and Pattern Recognition (CVPR)}, 2023.

\bibitem[Qi et~al.(2017{\natexlab{a}})Qi, Su, Mo, and Guibas]{Qi_2017_CVPR}
Charles~R. Qi, Hao Su, Kaichun Mo, and Leonidas~J. Guibas.
\newblock Pointnet: Deep learning on point sets for 3d classification and
  segmentation.
\newblock In \emph{Proceedings of the IEEE Conference on Computer Vision and
  Pattern Recognition (CVPR)}, 2017{\natexlab{a}}.

\bibitem[Qi et~al.(2017{\natexlab{b}})Qi, Yi, Su, and Guibas]{Qi_2017_NEURIPS}
Charles~Ruizhongtai Qi, Li Yi, Hao Su, and Leonidas~J Guibas.
\newblock Pointnet++: Deep hierarchical feature learning on point sets in a
  metric space.
\newblock In \emph{Advances in Neural Information Processing Systems}. Curran
  Associates, Inc., 2017{\natexlab{b}}.

\bibitem[Radford et~al.(2021)Radford, Kim, Hallacy, Ramesh, Goh, Agarwal,
  Sastry, Askell, Mishkin, Clark, et~al.]{radford_2021_PMLR}
Alec Radford, Jong~Wook Kim, Chris Hallacy, Aditya Ramesh, Gabriel Goh,
  Sandhini Agarwal, Girish Sastry, Amanda Askell, Pamela Mishkin, Jack Clark,
  et~al.
\newblock Learning transferable visual models from natural language
  supervision.
\newblock In \emph{International conference on machine learning}, pages
  8748--8763. PMLR, 2021.

\bibitem[Rozenberszki et~al.(2022)Rozenberszki, Litany, and
  Dai]{rozenberszki2022language}
David Rozenberszki, Or Litany, and Angela Dai.
\newblock Language-grounded indoor 3d semantic segmentation in the wild.
\newblock In \emph{Proceedings of the European Conference on Computer Vision
  (ECCV)}, pages 125--141. Springer, 2022.

\bibitem[Sanghi(2020)]{sanghi2020info3d}
Aditya Sanghi.
\newblock Info3d: Representation learning on 3d objects using mutual
  information maximization and contrastive learning.
\newblock In \emph{Computer Vision--ECCV 2020: 16th European Conference,
  Glasgow, UK, August 23--28, 2020, Proceedings, Part XXIX 16}, pages 626--642.
  Springer, 2020.

\bibitem[Sarkar et~al.(2023)Sarkar, Miksik, Pollefeys, Barath, and
  Armeni]{sarkar2023sgaligner}
Sayan~Deb Sarkar, Ondrej Miksik, Marc Pollefeys, Daniel Barath, and Iro Armeni.
\newblock Sgaligner : 3d scene alignment with scene graphs.
\newblock \emph{Proceedings of the IEEE International Conference on Computer
  Vision (ICCV)}, 2023.

\bibitem[Sauder and Sievers(2019)]{sauder2019self}
Jonathan Sauder and Bjarne Sievers.
\newblock Self-supervised deep learning on point clouds by reconstructing
  space.
\newblock \emph{Advances in Neural Information Processing Systems}, 32, 2019.

\bibitem[Schult et~al.(2023)Schult, Engelmann, Hermans, Litany, Tang, and
  Leibe]{Schult_2022_CORR}
Jonas Schult, Francis Engelmann, Alexander Hermans, Or Litany, Siyu Tang, and
  Bastian Leibe.
\newblock {Mask3D: Mask Transformer for 3D Semantic Instance Segmentation}.
\newblock 2023.

\bibitem[Tian et~al.(2020)Tian, Krishnan, and Isola]{tian2020contrastive}
Yonglong Tian, Dilip Krishnan, and Phillip Isola.
\newblock Contrastive multiview coding.
\newblock In \emph{Computer Vision--ECCV 2020: 16th European Conference,
  Glasgow, UK, August 23--28, 2020, Proceedings, Part XI 16}, pages 776--794.
  Springer, 2020.

\bibitem[Turc et~al.(2019)Turc, Chang, Lee, and Toutanova]{turc2019well}
Iulia Turc, Ming-Wei Chang, Kenton Lee, and Kristina Toutanova.
\newblock Well-read students learn better: On the importance of pre-training
  compact models.
\newblock \emph{arXiv preprint arXiv:1908.08962}, 2019.

\bibitem[Wald et~al.(2019)Wald, Avetisyan, Navab, Tombari, and
  Niessner]{Wald_2019_ICCV}
Johanna Wald, Armen Avetisyan, Nassir Navab, Federico Tombari, and Matthias
  Niessner.
\newblock Rio: 3d object instance re-localization in changing indoor
  environments.
\newblock In \emph{Proceedings of the IEEE/CVF International Conference on
  Computer Vision (ICCV)}, 2019.

\bibitem[Wald et~al.(2020)Wald, Dhamo, Navab, and Tombari]{Wald_2020_CVPR}
Johanna Wald, Helisa Dhamo, Nassir Navab, and Federico Tombari.
\newblock Learning 3d semantic scene graphs from 3d indoor reconstructions.
\newblock In \emph{Proceedings of the IEEE/CVF Conference on Computer Vision
  and Pattern Recognition (CVPR)}, 2020.

\bibitem[Wald et~al.(2022)Wald, Navab, and Tombari]{wald2022learning}
Johanna Wald, Nassir Navab, and Federico Tombari.
\newblock Learning 3d semantic scene graphs with instance embeddings.
\newblock \emph{International Journal of Computer Vision}, 130\penalty0
  (3):\penalty0 630--651, 2022.

\bibitem[Wang et~al.(2021)Wang, Liu, Yue, Lasenby, and Kusner]{Wang_2021_ICCV}
Hanchen Wang, Qi Liu, Xiangyu Yue, Joan Lasenby, and Matt~J. Kusner.
\newblock Unsupervised point cloud pre-training via occlusion completion.
\newblock In \emph{Proceedings of the IEEE/CVF International Conference on
  Computer Vision (ICCV)}, pages 9782--9792, 2021.

\bibitem[Wang et~al.(2023)Wang, Cheng, Zhao, Xu, Tang, and Sheng]{wang2023vl}
Ziqin Wang, Bowen Cheng, Lichen Zhao, Dong Xu, Yang Tang, and Lu Sheng.
\newblock Vl-sat: Visual-linguistic semantics assisted training for 3d semantic
  scene graph prediction in point cloud.
\newblock \emph{arXiv preprint arXiv:2303.14408}, 2023.

\bibitem[Wu et~al.(2021)Wu, Wald, Tateno, Navab, and Tombari]{Wu_2021_CVPR}
Shun-Cheng Wu, Johanna Wald, Keisuke Tateno, Nassir Navab, and Federico
  Tombari.
\newblock Scenegraphfusion: Incremental 3d scene graph prediction from rgb-d
  sequences.
\newblock In \emph{Proceedings of the IEEE/CVF Conference on Computer Vision
  and Pattern Recognition (CVPR)}, pages 7515--7525, 2021.

\bibitem[Wu et~al.(2023)Wu, Tateno, Navab, and Tombari]{Wu_2023_CVPR}
Shun-Cheng Wu, Keisuke Tateno, Nassir Navab, and Federico Tombari.
\newblock Incremental 3d semantic scene graph prediction from rgb sequences.
\newblock In \emph{Proceedings of the IEEE/CVF Conference on Computer Vision
  and Pattern Recognition (CVPR)}, pages 5064--5074, 2023.

\bibitem[Wu et~al.(2015)Wu, Song, Khosla, Yu, Zhang, Tang, and
  Xiao]{Wu_2015_CVPR}
Zhirong Wu, Shuran Song, Aditya Khosla, Fisher Yu, Linguang Zhang, Xiaoou Tang,
  and Jianxiong Xiao.
\newblock 3d shapenets: A deep representation for volumetric shapes.
\newblock In \emph{Proceedings of the IEEE Conference on Computer Vision and
  Pattern Recognition (CVPR)}, 2015.

\bibitem[Wu et~al.(2018)Wu, Xiong, Yu, and Lin]{Wu_2018_CVPR}
Zhirong Wu, Yuanjun Xiong, Stella~X. Yu, and Dahua Lin.
\newblock Unsupervised feature learning via non-parametric instance
  discrimination.
\newblock In \emph{Proceedings of the IEEE Conference on Computer Vision and
  Pattern Recognition (CVPR)}, 2018.

\bibitem[Xie et~al.(2020)Xie, Gu, Guo, Qi, Guibas, and Litany]{Xie_2020_ECCV}
Saining Xie, Jiatao Gu, Demi Guo, Charles~R. Qi, Leonidas Guibas, and Or
  Litany.
\newblock Pointcontrast: Unsupervised pre-training for 3d point cloud
  understanding.
\newblock In \emph{Computer Vision -- ECCV 2020}, pages 574--591, Cham, 2020.
  Springer International Publishing.

\bibitem[Xu et~al.(2017)Xu, Zhu, Choy, and Fei-Fei]{Xu_2017_CVPR}
Danfei Xu, Yuke Zhu, Christopher~B. Choy, and Li Fei-Fei.
\newblock Scene graph generation by iterative message passing.
\newblock In \emph{Proceedings of the IEEE Conference on Computer Vision and
  Pattern Recognition (CVPR)}, 2017.

\bibitem[Xue et~al.(2023{\natexlab{a}})Xue, Gao, Xing, Mart{\'\i}n-Mart{\'\i}n,
  Wu, Xiong, Xu, Niebles, and Savarese]{Xue_2023_CVPR}
Le Xue, Mingfei Gao, Chen Xing, Roberto Mart{\'\i}n-Mart{\'\i}n, Jiajun Wu,
  Caiming Xiong, Ran Xu, Juan~Carlos Niebles, and Silvio Savarese.
\newblock Ulip: Learning a unified representation of language, images, and
  point clouds for 3d understanding.
\newblock In \emph{Proceedings of the IEEE/CVF Conference on Computer Vision
  and Pattern Recognition (CVPR)}, pages 1179--1189, 2023{\natexlab{a}}.

\bibitem[Xue et~al.(2023{\natexlab{b}})Xue, Yu, Zhang, Li, Martín-Martín, Wu,
  Xiong, Xu, Niebles, and Savarese]{xue2023ulip2}
Le Xue, Ning Yu, Shu Zhang, Junnan Li, Roberto Martín-Martín, Jiajun Wu,
  Caiming Xiong, Ran Xu, Juan~Carlos Niebles, and Silvio Savarese.
\newblock Ulip-2: Towards scalable multimodal pre-training for 3d
  understanding, 2023{\natexlab{b}}.

\bibitem[Yang et~al.(2018)Yang, Lu, Lee, Batra, and Parikh]{Yang_2018_ECCV}
Jianwei Yang, Jiasen Lu, Stefan Lee, Dhruv Batra, and Devi Parikh.
\newblock Graph r-cnn for scene graph generation.
\newblock In \emph{Proceedings of the European Conference on Computer Vision
  (ECCV)}, 2018.

\bibitem[Yuksekgonul et~al.(2023)Yuksekgonul, Bianchi, Kalluri, Jurafsky, and
  Zou]{yuksekgonul2023when}
Mert Yuksekgonul, Federico Bianchi, Pratyusha Kalluri, Dan Jurafsky, and James
  Zou.
\newblock When and why vision-language models behave like bags-of-words, and
  what to do about it?
\newblock In \emph{International Conference on Learning Representations}, 2023.

\bibitem[Zhang et~al.(2021{\natexlab{a}})Zhang, Yu, Song, and
  Cai]{Zhang_2021_CVPR}
Chaoyi Zhang, Jianhui Yu, Yang Song, and Weidong Cai.
\newblock Exploiting edge-oriented reasoning for 3d point-based scene graph
  analysis.
\newblock In \emph{Proceedings of the IEEE/CVF Conference on Computer Vision
  and Pattern Recognition (CVPR)}, pages 9705--9715, 2021{\natexlab{a}}.

\bibitem[Zhang et~al.(2022)Zhang, Zhang, Hu, Chen, Li, Dai, Wang, Yuan, Hwang,
  and Gao]{zhang2022glipv2}
Haotian* Zhang, Pengchuan* Zhang, Xiaowei Hu, Yen-Chun Chen, Liunian~Harold Li,
  Xiyang Dai, Lijuan Wang, Lu Yuan, Jenq-Neng Hwang, and Jianfeng Gao.
\newblock Glipv2: Unifying localization and vision-language understanding.
\newblock \emph{arXiv preprint arXiv:2206.05836}, 2022.

\bibitem[Zhang et~al.(2021{\natexlab{b}})Zhang, Girdhar, Joulin, and
  Misra]{Zhang_2021_ICCV}
Zaiwei Zhang, Rohit Girdhar, Armand Joulin, and Ishan Misra.
\newblock Self-supervised pretraining of 3d features on any point-cloud.
\newblock In \emph{Proceedings of the IEEE/CVF International Conference on
  Computer Vision (ICCV)}, pages 10252--10263, 2021{\natexlab{b}}.

\end{thebibliography}
}
\setcounter{page}{1}
\setcounter{figure}{5}
\setcounter{table}{5}
\setcounter{equation}{9}
\maketitlesupplementary

This document supplements our work \textit{Lang3DSG: Language-based contrastive pre-training for 3D scene graph prediction} by providing 
\begin{enumerate*}[label=\enumlabel]
    \item reproducibility information on our implementation and architecture (\cref{sec:reproducibility}),
    \item details on the importance of negative samples during training (\cref{sec:negatives}),
    \item investigations into the understanding capabilities of CLIP for compositional scene descriptions (\cref{sec:composition}),
    \item additional 3D scene graph generations from diverse scenes (\cref{sec:sg_viz}),
    \item additional examples of zero-shot room type classification using our language-aligned features (\cref{sec:sg_zeroshot}),
\end{enumerate*}

\section{Reproducibility}
\label{sec:reproducibility}
Our encoder consists of two PointNets which pass features of size 256 to a 4-layer GCN, where g1(·) and g2(·) are composed of a linear layer followed by a ReLU activation. The projectors are 3-layer MLPs with ReLU activation in the first two layers and feature dimensions of $[256,1024,512]$ for CLIP ViT-B/32 and BERT and $[256,1024,768]$ for CLIP ViT-V/14. 
During fine-tuning, we replace the projectors with object and predicate prediction MLPs consisting of 3 linear layers with feature dimensions $[256,512,class\_num]$ with batch normalization and ReLU activation. 

The training is performed on 1 NVIDIA A100 GPU with 80 GB memory.

\section{Role of negatives during pre-training}
\label{sec:negatives}
\begin{figure}[t]
    \centering
    \begin{subfigure}{0.45\linewidth}
    \includegraphics[width=0.99\textwidth]{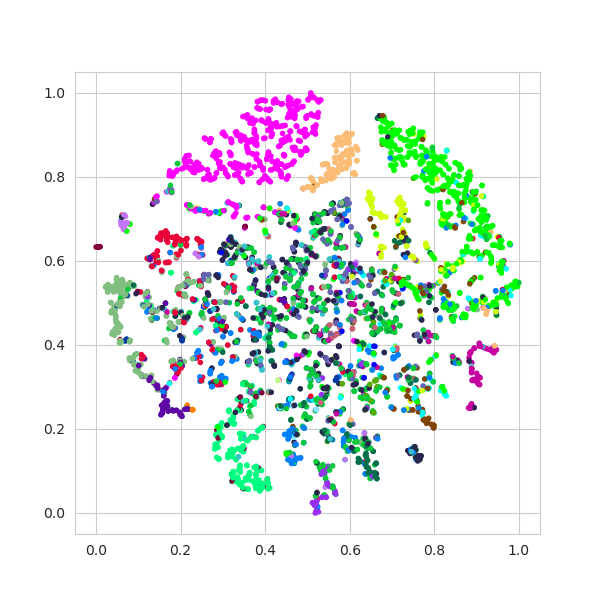}
    \caption{Objects}
    \end{subfigure}
    \begin{subfigure}{0.45\linewidth}
    \includegraphics[width=0.99\textwidth]{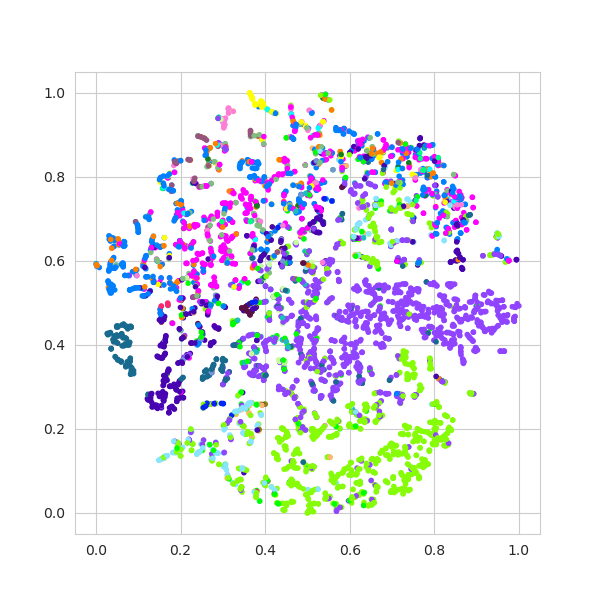}
    \caption{Predicates}
    \end{subfigure}\\
    w/o negatives\\
    \begin{subfigure}{0.45\linewidth}
    \includegraphics[width=0.99\textwidth]{images/objects_pretrain.png}
    \caption{Objects}
    \end{subfigure}
    \begin{subfigure}{0.45\linewidth}
    \includegraphics[width=0.99\textwidth]{images/predicates_pretrain_new.png}
    \caption{Predicates}
    \end{subfigure}\\
     16 negatives
    \caption{\textbf{t-SNE embedding for pre-training w/ and w/o negatives.} We observe better clustering for objects and predicates using negatives.}
    \label{fig:tsne_nonegatives}
\end{figure}

In Sec. 3 of the main paper, we describe our contrastive pre-training. We use a cosine similarity loss to distill the knowledge of CLIP into our 3D graph model. For this, we differentiate between positive cases where our goal is to maximize the cosine similarity (see Eq.~6) and negative cases where our goal is to minimize the cosine similarity (see Eq.~7). 
This formulation is adapted from classical contrastive representation learning, where negative samples are needed to prevent a collapse of the latent representation. However, since we are trying to distill the knowledge directly from CLIP, negative samples are not strictly necessary. 
However, in \cref{fig:tsne_nonegatives}, we show the difference in the learned latent embedding with and without negatives. 
It is important to note that both training with negatives and without negatives produce a structured latent embedding. 
However, we find that the latent representation for individual classes trained without negatives is much more mixed. This mix of class embeddings leads to a reduced effect of our pre-training which can be seen in \cref{tab:wo_negatives_vs_with}.

\begin{table}[t]
\tabcolsep=3.4mm
\centering
\small
\begin{tabular}{@{}lcccc@{}}
\toprule
       & \multicolumn{2}{c}{Object} & \multicolumn{2}{c}{Predicate}\\
       \cmidrule(lr){2-3}\cmidrule(lr){4-5}
       num negatives & R@5                 & mR@5                & R@3                              & mR@3                              \\
      \midrule
      0  & 0.74 & 0.40 & 0.95 & 0.65 \\
      16  & 0.77 & 0.43 & 0.96 & 0.67 \\
      32  & 0.76 & 0.42 & 0.96 & 0.66 \\
\bottomrule
\end{tabular}
\caption{\textbf{Role of negatives.} Using negatives during pre-training produces better fine-tuned results. The best results are achieved using 16 negatives. }
\label{tab:wo_negatives_vs_with}
\end{table}

\begin{figure*}[t]
    \centering
    \includegraphics[width=0.98\textwidth]{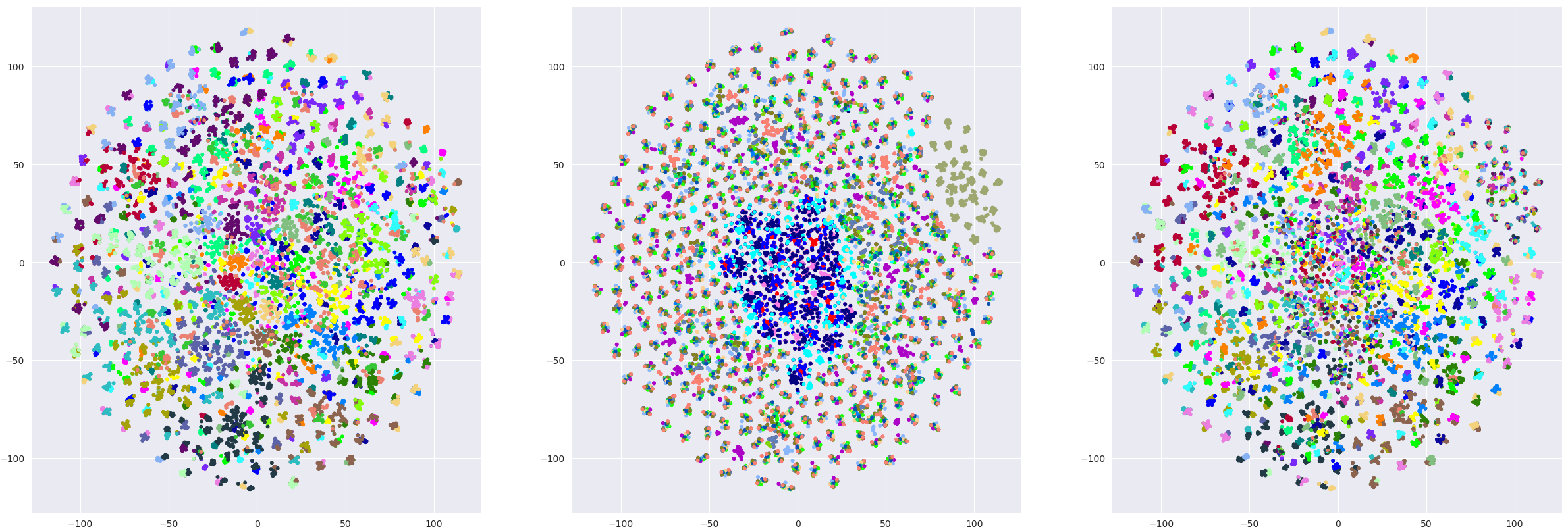}
    \begin{subfigure}{0.327\textwidth}
        \caption{Highlight subject}
    \end{subfigure}
    \begin{subfigure}{0.327\textwidth}
        \caption{Highlight predicate}
    \end{subfigure}\begin{subfigure}{0.327\textwidth}
        \caption{Highlight object}
    \end{subfigure}
    \caption{\textbf{CLIP embedding space for complex relationships.} We provide the same t-SNE projection for the same CLIP embeddings, highlighted by subject in (a), by predicate (b) and by object in (c). We observe that the embedding mostly clusters by subject. Some clustering can be observed for the object in the relationship. However, CLIP appears to attend less to the predicate in the relationship since no clustering for multiple classes can be observed.}
    \label{fig:tsne_clip}
\end{figure*}

\section{CLIP relationship understanding}
\label{sec:composition}
In this section, we investigate the relationship understanding capability of CLIP, raised from our findings in Tab. 4 from the main paper. A couple of works have investigated the understanding capabilities of CLIP for compositional scenes [13, 38, 65]. In contrast to these works which study the vision-language understanding of CLIP, our focus lies on language understanding only. 
The applicability of their investigations is therefore uncertain for our approach. However, our experiments in Tab. 4 indicate that CLIP is only able to extract limited knowledge from provided relationship descriptions of the form "A scene of a [subject] is [predicate] a [object]". 
To validate this issue, we perform an additional experiment visualizing the generated embeddings with CLIP in \cref{fig:tsne_clip}. 
We plot the embeddings for all combinations for subject, predicate and objects from a subsampled set of 27 object classes and 27 predicate classes resulting in $27^3=19683$ unique relationships. We embed these relationships with CLIP and project their features using t-SNE. 
In \cref{fig:tsne_clip} we provide this projection three times, with colored coding for the subject in (a), color coding for the predicate in (b), and color coding for the object in the relationship in (c). We observe that the feature projections generally cluster together by subject, with some clustering also appearing for objects. 
However, there does not appear to be particular clustering for predicates. This indicates that the investigations about compositionality from [13, 38, 65] also apply to the language-only part of CLIP. 
We see the potential for future research in training CLIP-like models to understand complex relationships.

\section{Additional 3D scene graph predictions}
\label{sec:sg_viz}
Here we provide additional 3D scene graph predictions for a diverse set of 3D scenes. Overall, these examples confirm the results from the main paper. Most nodes and edges are predicted correctly with some edges being predicted partially correctly and only a very few nodes and edges are misclassified.

\section{Additional zero-shot room type predictions}
\label{sec:sg_zeroshot}
Here we provide additional qualitative examples for our proposed zero-shot room type classification. 
We provide different 3D scenes with their softmax probability based on the similarity scores between the pooled 3D graph feature and the text queries. 
We analyze the zero-shot capabilities on 3RScan, which consists of mostly indoor home scenes. 
We, therefore, choose to evaluate the room types "bathroom", "dining area", "kitchen" and "living room". There are no room-based labels in the datasets, therefore we evaluate the room type prediction on a qualitative basis only. We observe that the zero-shot predictions are very accurate and with high confidence in most of the cases. 
However, we notice that scenes representing dining areas get misclassified as living rooms. 
In this experiment, we note that the similarity between the text embedding of "dining area" and our language-aligned 3D graph features is generally low, resulting in a small softmax score for all scenes. For the dining area scenes, however, this similarity score is considerably higher but gets outscored by text queries with a stronger response.

\begin{figure*}[t]
    \centering
    \includegraphics[width=1.0\textwidth]{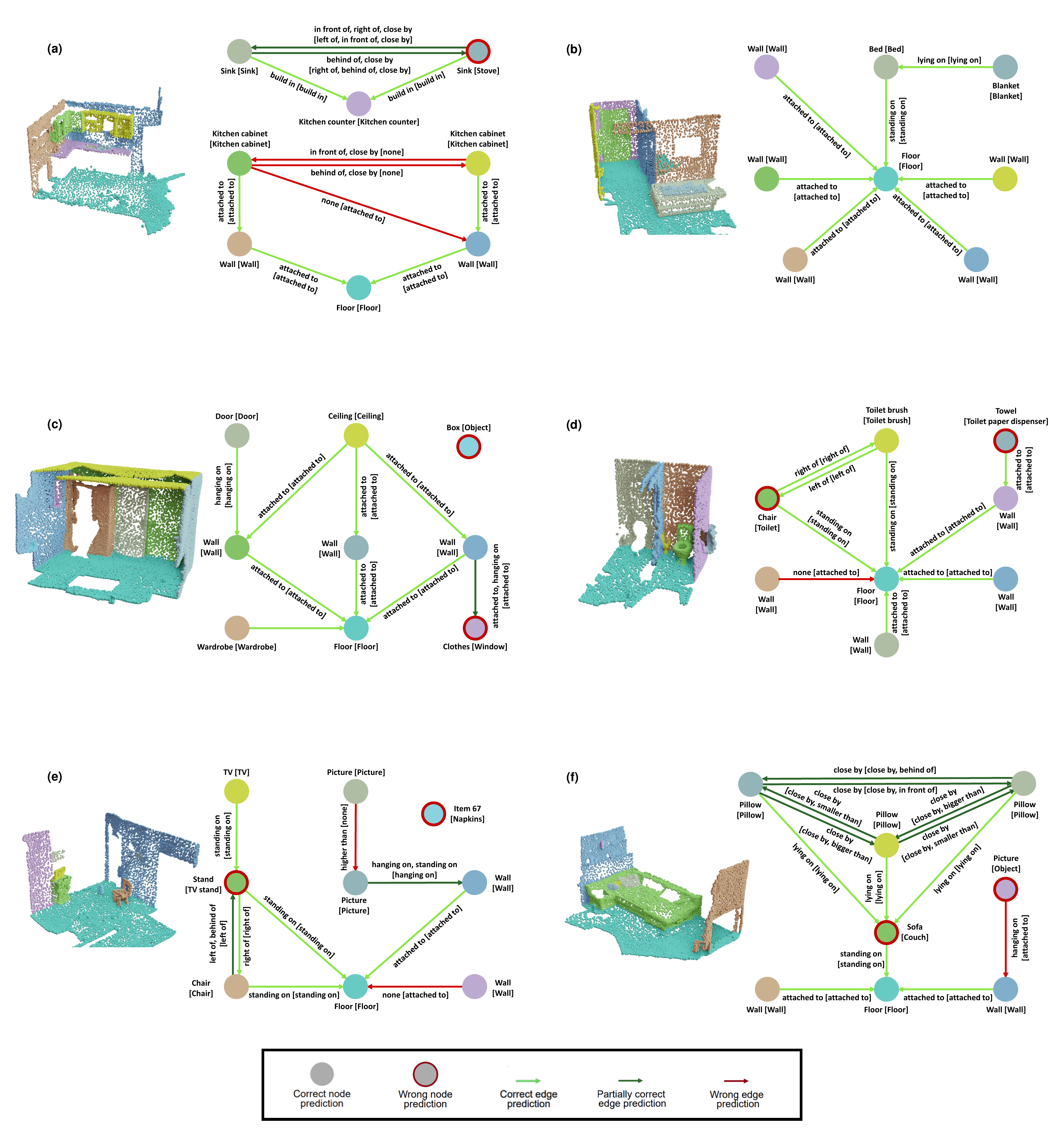}
    \caption{\textbf{3D scene graph predictions.}}
    \label{fig:supp_graphs}
\end{figure*}

\begin{figure*}[t]
    \centering
    \includegraphics[height=0.95\textheight]{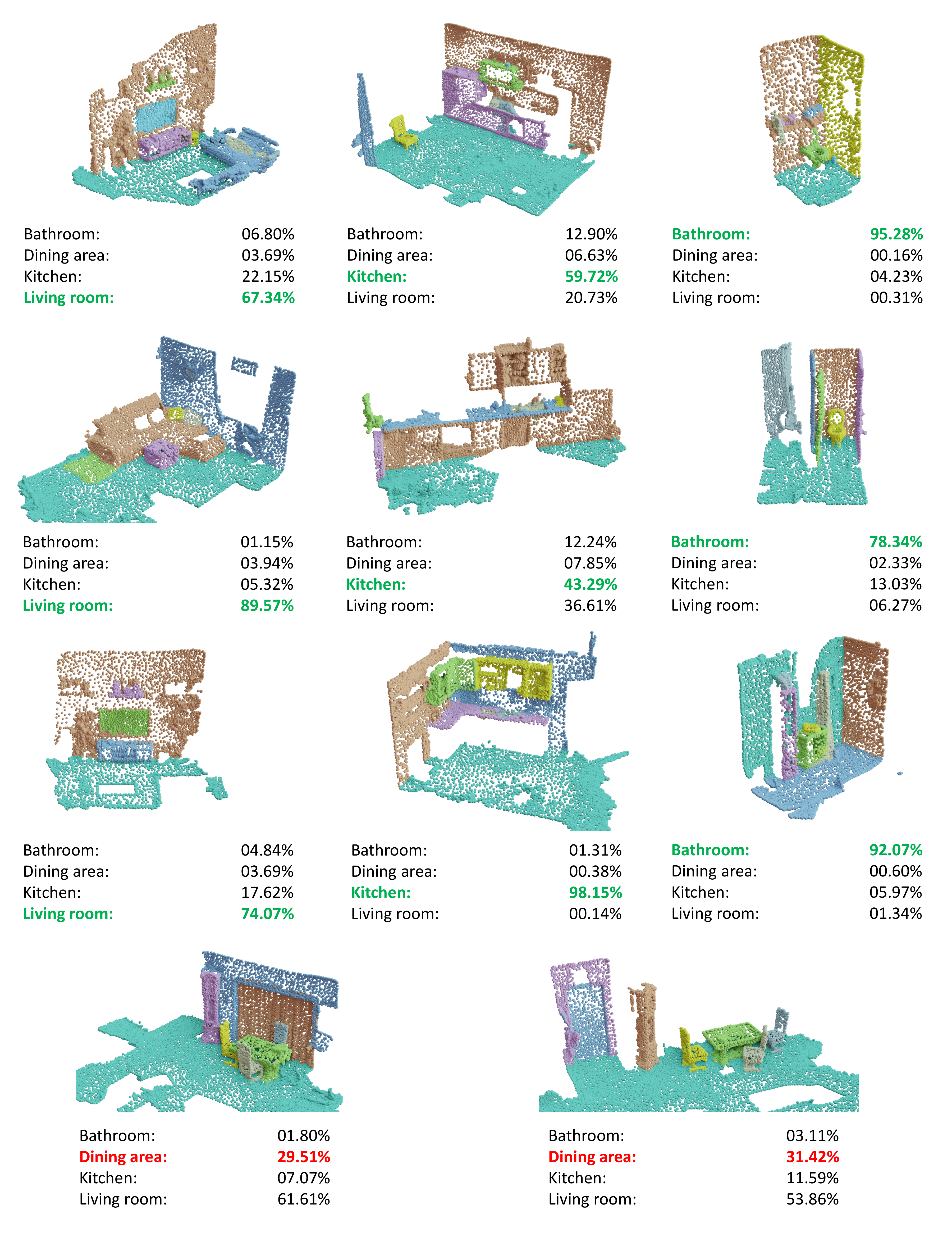}
    \caption{\textbf{Zero-shot room type predictions.}}
    \label{fig:supp_zeroshot}
\end{figure*}

\end{document}